\documentclass[journal]{IEEEtran}
\usepackage{}
\usepackage{mathrsfs}
\usepackage{color}

\ifCLASSINFOpdf
\else
\fi
\usepackage{graphics}
\usepackage{epsfig}
\usepackage[ruled]{algorithm2e}
\usepackage{booktabs}
\usepackage{url}
\usepackage{psfrag}

\usepackage{amsfonts}
\usepackage{amsmath, amssymb}
\usepackage[tight,footnotesize]{subfigure}

\usepackage{epstopdf}

\newcommand{\reals}{{\mathbb{R}}}

\newcommand{\cL}{{\cal L}}

\newcommand{\sS}{{\mathbb{S}}}

\newcommand{\vx}{{\bf x}}
\newcommand{\vs}{{\bf s}}

\newcommand{\ve}{{\bf e}}
\newcommand{\vu}{{\bf u}}

\newcommand{\vz}{{\bf z}}

\newcommand{\vp}{{\bf p}}
\newcommand{\vq}{{\bf q}}

\newcommand{\vt}{{\bf t}}

\newcommand{\vzero}{{\bf 0}}

\newcommand{\vU}{{\bf U}}

\newcommand{\valpha}{\boldsymbol{\alpha}}
\newcommand{\vbeta}{\boldsymbol{\beta}}

\begin{document}
%
\title{Mixed One-bit Compressive Sensing with Application to Overexposure Correction for CT Reconstruction}
%
%
%

\author{Xiaolin Huang,~\IEEEmembership{Member,~IEEE},
        Yan Xia, Lei~Shi, Yixing Huang, \\ Ming~Yan, Joachim~Hornegger,
        and~Andreas Maier,~\IEEEmembership{Member,~IEEE}
\thanks{

This work is partially supported by Alexander von Humboldt Foundation and National Natural Science Foundation of China (NSFC, 61603248). L. Shi is supported by the NSFC (11571078), the Joint Research Fund by NSFC and Research Grants Council of Hong Kong (11461161006 and CityU 104012) and ``Zhuo Xue'' program of Fudan University.

X. Huang is with Institute of Image Processing and Pattern Recognition, Shanghai Jiao Tong University, Shanghai, P.R. China.
Y. Xia is with Department of Radiology, Stanford University, CA, USA.
L. Shi is with School of Mathematical Sciences, Fudan University, Shanghai, P.R. China.
Y. Huang, J. Hornegger, and A. Maier are with Pattern Recognition Lab of the Friedrich-Alexander-Universit\"{a}t Erlangen-N\"{u}rnberg, Erlangen, Germany. M. Yan (corresponding author) is with the Department of Computational Mathematics, Science and Engineering, Michigan State University, MI, USA. (e-mails: xiaolinhuang@sjtu.edu.cn, yaxia@stanford.edu, leishi@fudan.edu.cn, yixing.huang@fau.de, yanm@math.msu.edu, joachim.hornegger@fau.de, andreas.maier@fau.de.)}
}

%
%

\markboth{}%
{Huang \MakeLowercase{\textit{et al.}}: Mixed 1bit-CS with Applications on CT Reconstruction}
%



\maketitle

\begin{abstract}
When a measurement falls outside the quantization or measurable range, it becomes saturated and cannot be used in classical reconstruction methods.
For example, in C-arm angiography systems, which provide projection radiography,
fluoroscopy, digital subtraction angiography, and are widely used for medical
diagnoses and interventions, the limited dynamic range of
C-arm flat detectors leads to overexposure in some projections during an acquisition, such as imaging relatively thin body parts
(e.g., the knee). Aiming at overexposure correction for computed tomography (CT) reconstruction, we in this paper propose a mixed one-bit compressive sensing (M1bit-CS) to acquire information from both regular and saturated measurements. This method is inspired by the recent progress on one-bit compressive sensing, which deals with only sign observations. Its successful applications imply that information carried by saturated measurements is useful to improve recovery quality. For the proposed M1bit-CS model, alternating direction methods of multipliers is developed and an iterative saturation detection scheme is established. Then we evaluate M1bit-CS on one-dimensional signal recovery tasks. In some experiments, the performance of the proposed algorithms on mixed measurements is almost the same as recovery on unsaturated ones with the same amount of measurements. Finally, we apply the proposed method to overexposure correction for CT reconstruction on a phantom and a simulated clinical image. The results are promising, as the typical streaking
artifacts and capping artifacts introduced by saturated projection data are effectively reduced, yielding significant error reduction compared with existing algorithms based on extrapolation.
\end{abstract}



%
\IEEEpeerreviewmaketitle

\section{Introduction}
\IEEEPARstart{I}{n} a real-world measuring system, the measurable range could be limited and the final output for a linear measurement can be described as
\begin{equation}\label{saturation-data}
 p_i = \max \Big \{s^-, \min \big\{\vu_i^\top  \vx + \varepsilon, s^+ \big\} \Big \},
\end{equation}
where $\vx \in \reals^d$ is the signal to be estimated, $\vu_i \in \reals^d$ is a sensing vector,  $\varepsilon$ stands for the noise, and $s^-$ and $s^+$ are the lower and upper saturation thresholds, respectively. When an observation $p_i$ equals $s^+$ or $s^-$, we say that this measurement is \emph{saturated}. The saturation phenomenon may happen when the detector has a narrow range~\cite{feng1998sensor,fang2013afm,dos2013ct}. In this paper, we discuss how to recover $\vx$ from a sensing system with saturation and then apply the proposed method to correct overexposure for C-arm computed tomography (C-arm CT).

C-arm CT is characterized with high spatial resolution, large field of view, and 3D imaging
capability. It has become a valuable tool for medical diagnoses, therapy
guidance, and interventions. The key technology that enables these
advances is the flat-panel detector, which provides good soft-tissue imaging
performance, distortion-free images as well as high detective quantum
efficiency \cite{Strobel08:IWF}. However, although typical
C-arm flat-panel detectors have a dynamic range of 14-bit that is
sufficient for conventional fluoroscopy, it may not be high enough
to eliminate overexposure in all projections acquired during a 3D acquisition, such as imaging relatively thin body parts (e.g., the
knee). If no counter-measures are carried out, strong direct radiation
may hit certain detector regions in some views (the intensity range
of the traveled X-rays are greater than the detector's inherent detectable
range) and thus overexpose them. Consequently, reconstruction from
overexposed projections typically leads to incorrect Hounsfield unit
(HU) values, streaking artifacts, and loss of clear outer boundaries.
Moreover, HU values of a homogeneous object tend to become smaller
when they are far away from the object center, resulting in so-called
capping artifacts. Since all these overexposure artifacts impair the
final visualization of low contrast objects, it is of practical significance
to develop an appropriate overexposure correction scheme on the saturated
images to compensate for these artifacts.

\begin{figure}[htbp]
  \centering
  \subfigure[]{\label{fig-phantom:a}
    \includegraphics[width=0.4\linewidth]{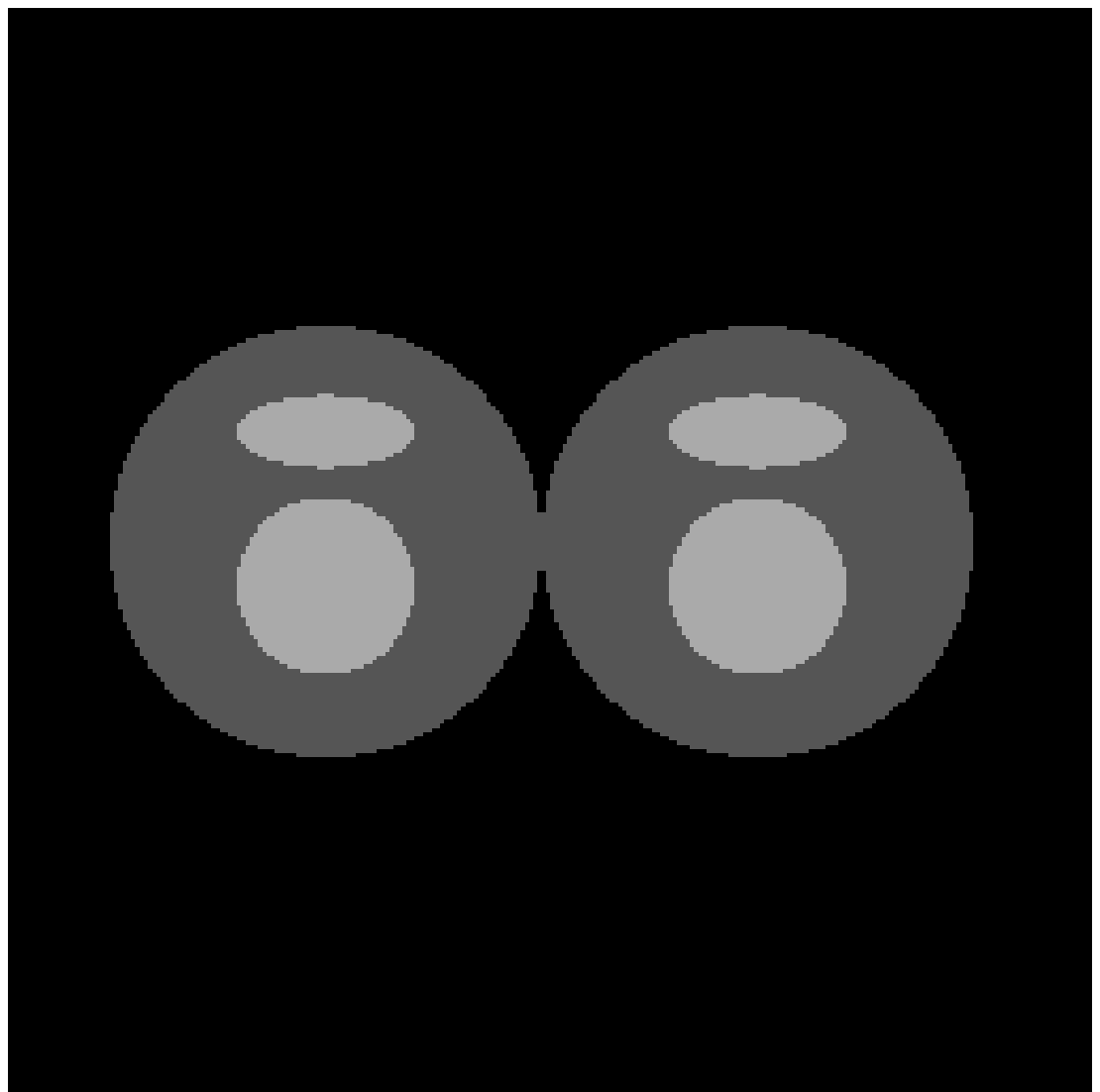}} \quad
  \subfigure[]{\label{fig-phantom:b}
    \includegraphics[width=0.4\linewidth]{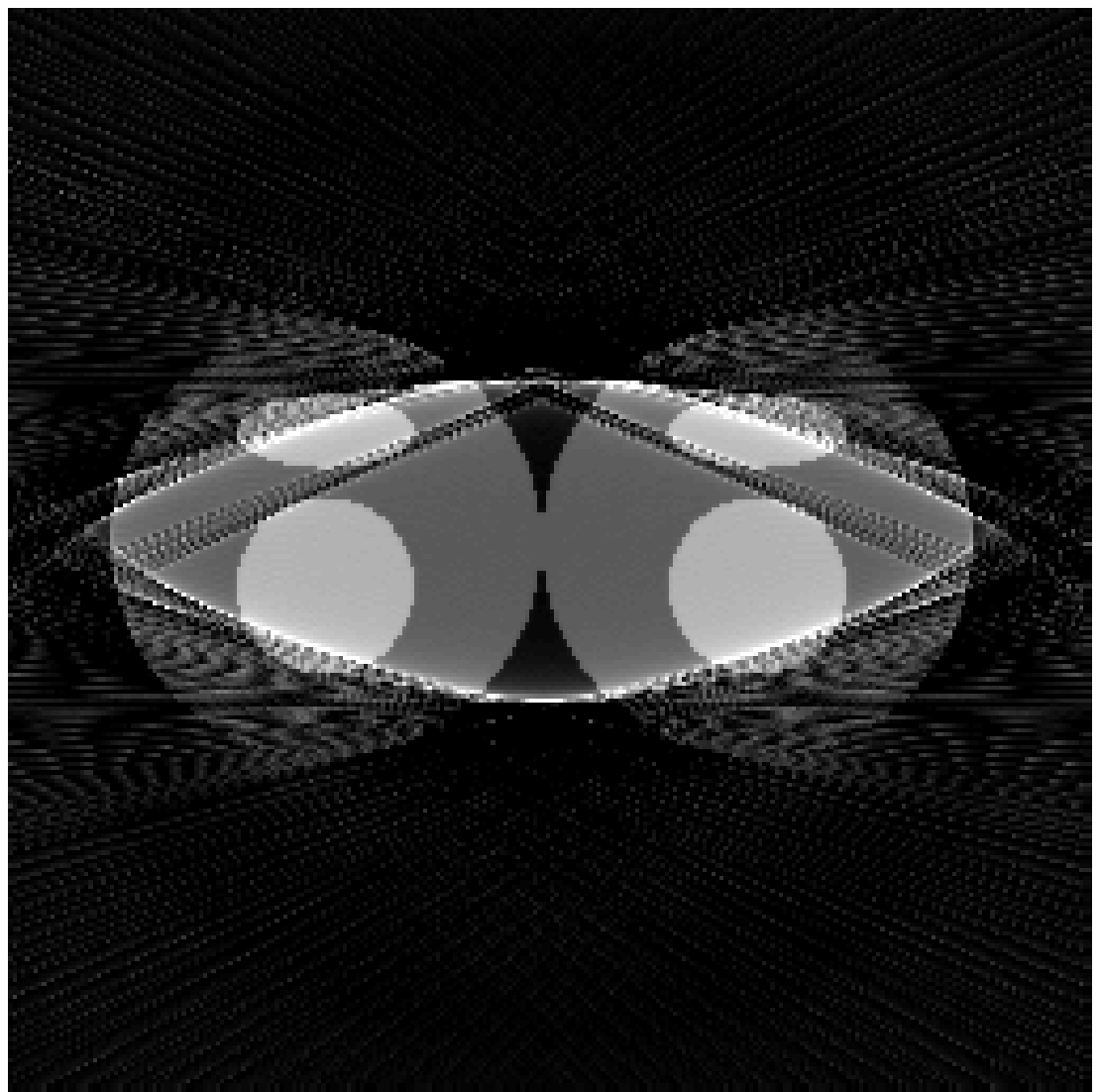}}\\
    \subfigure[]{\label{fig-phantom-projection:a}
    \includegraphics[width=0.77\linewidth]{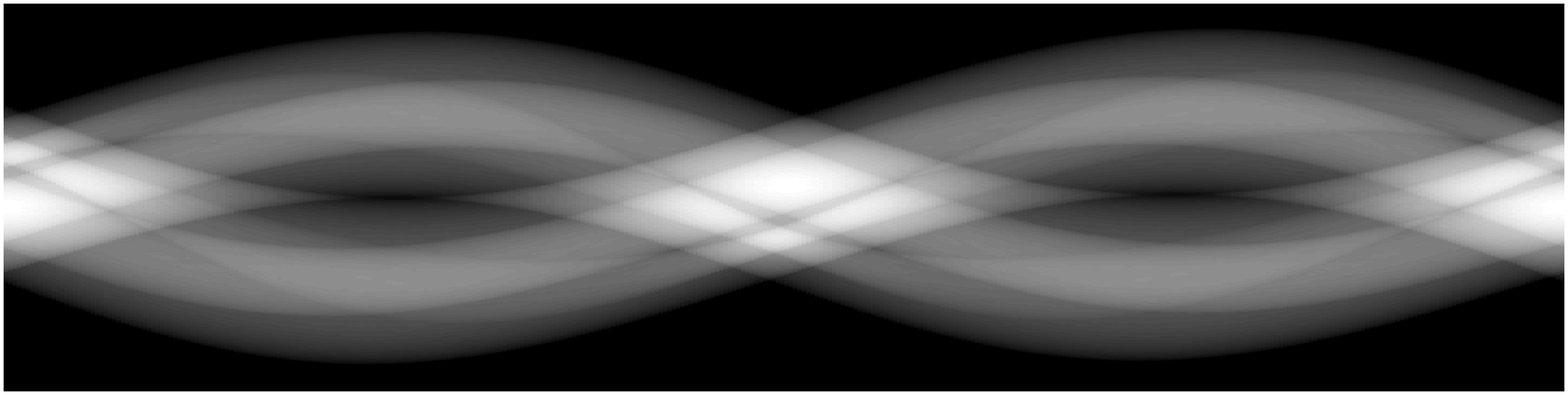}}\\
  \subfigure[]{\label{fig-phantom-projection:b}
    \includegraphics[width=0.77\linewidth]{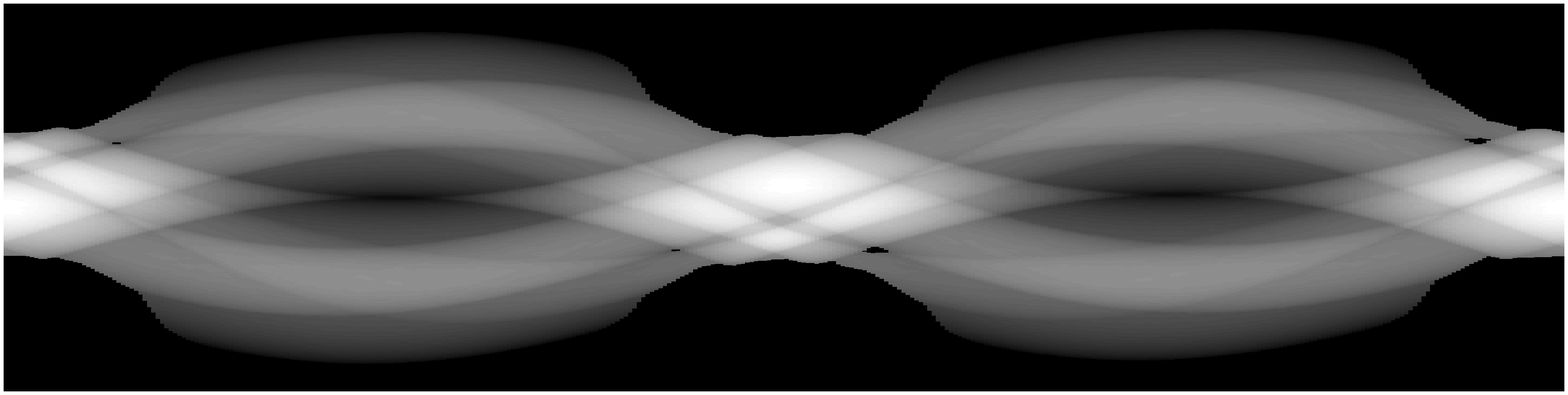}}
  \caption{(a) A knee phantom; (b) reconstructed by FBP from saturated projections shown by (d); (c) full projection data; (d) saturated projection data (the measurable range $\kappa = 0.5 p_{\max}$; see Section \ref{sec:CT-problem} for details).}\label{fig-phantom}
\end{figure}

To give an intuitive impression of saturation phenomenon happens in CT scan, a knee phantom is designed in Fig.\ref{fig-phantom:a}. Its $360^\circ$ projection data without and with saturation are shown in Fig.\ref{fig-phantom-projection:a} and \ref{fig-phantom-projection:b}, respectively. If there is no saturation, then the object can be reconstructed well by classical analytical algorithms, such as the standard filtered back projection (FBP, \cite{feldkamp1984practical}) or iterative reconstruction methods, such as the simultaneous algebraic reconstruction technique (SART,~\cite{andersen1984simultaneous}). When there is saturation in the sinogram data, e.g. Fig.1(d), CT reconstruction from the saturated data is quite hard. For example, directly applying FBP on the saturated data in Fig.\ref{fig-phantom-projection:b} outputs Fig.\ref{fig-phantom:b}, which is far from satisfactory.


Due to the loss of information, signal recovery from saturated measurements are not straightforward. In CT reconstruction, primarily targeted to correct the overexposure artifacts for knee imaging, an optimization-based multiple shape fitting approach was
suggested to extrapolate the missing data using a pair of cylinder shapes that are fitted in the sinogram domain~\cite{univis91587769}.
The parameters describing the cylinders are estimated from the overexposed data by minimizing the least square error. However, the method would
face challenges when it comes to complicated shapes that cannot be modeled with a composition of cylindrical or elliptical
shapes. Furthermore, additional hardware, such as a shaped pre-filteration (bowtie) can be deployed to correct the saturation during the data
acquisition \cite{mail2009influence}. However, that method requires the imaged object to be exactly centered in the rotation center and the use
of a bowtie filter also appears not flexible enough for the applications of C-arm CT systems. Another way to prevent saturation at the object is to apply modeling
clay wrapped around the imaged object as an additional absorber during the data acquisition \cite{choi2013fiducial,choi2014fiducial}. However,
the method is not ideal since the modeling clay may cause great discomfort for the patient. Alternatively, a Kinect camera was used to obtain
the object depth information, which can be combined into a C-arm system for correcting overexposured projections \cite{rausch2016kinect}.
More specifically, the Kinect depth data is used to find the points of intersection between the X-ray beam path and object surface. Overexposed
pixels on the projection images are corrected by extrapolating the absorption along the corresponding line integrals.

In contrast to the existing saturation correction methods that require prior-knowledge, in this paper we propose a data-driven method to find $\vx$ directly from measurements observed by (\ref{saturation-data}). The key is to acquire information from the saturated measurements, which do not provide analog information but do contain one-bit information: when a measurement is upper-saturated (lower-saturated), i.e.,  $p_i = s^+ (s^-)$, we know that $\vu_i^\top  \vx + \varepsilon$ is greater (smaller) than or equal to $s^+$ ($s^-$).

Data mining from inequalities is theoretically possible when $\vx$ is sparse, which has been verified by the recent progress on \emph{one-bit compressive sensing} (\emph{1bit-CS}).  The concept of 1bit-CS was firstly introduced by \cite{boufounos20081} to recover sparse signals from measurements containing only one-bit information, i.e., sign measurements. It is rooted in {\emph{compressive sensing}} (CS, \cite{candes2005decoding,donoho2006compressed,candes2006robust}), and there are some interesting discussions on both theory and algorithms; see, e.g., \cite{laska2011trust} \cite{yan2012robust} \cite{jacques2013robust} \cite{plan2013robust}  \cite{zhang2014efficient} \cite{zhu2015towards}. Unlike the pure 1bit-CS, (\ref{saturation-data}) has analog measurements as well. Therefore, we refer to the method dealing with (\ref{saturation-data}) as \emph{mixed one-bit compressive sensing (M1bit-CS)}. There is few study on this topic. To the best of our knowledge, it has been considered only by \cite{laska2011democracy,liu2014robust}. But they put major emphasis on quantization error and assumed that there was no noise on the saturation measurements.

In this paper, we are going to recover sparse signals from noise-corrupted measurements with saturation. The major contributions include:
\begin{itemize}
\item propose M1bit-CS models;

\item establish the corresponding alternating direction method of multipliers (ADMM);

\item design an iterative saturation detection scheme;

\item evaluate M1bit-CS on one-dimensional signal recovery problems;

\item apply M1bit-CS on overexpose correction for C-arm CT and achieve very promising performance.
\end{itemize}


Before discussing M1bit-CS, we introduce some notations to describe (\ref{saturation-data}) that is a linear sensing system with saturation. The dimension of the signal is denoted by $d$, the number of total measurements by $m$, and the number of saturated measurements by $n$. A boolean vector $\Psi$
is used to indicate a measurement is saturated or not. Mathematically,
\[
\Psi_i = 1 ~~\Leftrightarrow~~ \vu_i^\top  \vx + \varepsilon > s^+ \mathrm{~or~} \vu_i^\top  \vx + \varepsilon < s^-.
\]
Furthermore, for saturated measurements, i.e., $\Psi_i = 1$, we use $y_i \in \{-1, +1\}$ to distinguish the upper-saturation ($y_i = 1$) and lower-saturation ($y_i = -1$):
\begin{equation}\label{sat-inequality}
y_i(\vu_i^\top  \vx + \varepsilon - s_i) \geq 0, ~~\forall i: \Psi_i = 1,
\end{equation}
where $s_i = s^+, \forall i: y_i = 1$ and $s_i = s^-, \forall i: y_i = -1$. Then our task of signal recovery from (\ref{saturation-data}) is to find a sparse signal $\vx$ such that $\vu_i^\top \vx$ is close to $p_i$ when $\Psi_i = 0$ and coincides with the saturation inequality (\ref{sat-inequality}) when $\Psi_i = 1$.

The rest of this paper is organized as follows. In Section~\ref{sec:mix}, M1bit-CS models are established. Fast algorithms are designed in Section~\ref{sec:ADMM}. Section~\ref{sec:numerical} evaluates the proposed algorithms in numerical experiments. Section~\ref{sec:CT} applies the proposed method to overexposure correction for C-arm CT reconstruction. In Section \ref{sec:conclusion}, a conclusion is given to end this paper.

\section{Mixed One-bit Compressive Sensing}\label{sec:mix}

\subsection{(One-bit) Compressive Sensing}
A classical compressive sensing (CS) model for unsaturated data, i.e., $\Psi_i = 0, \forall i$, can be formulated as the following well-known model:
\begin{align}\label{CS}
\min_{\vx \in \reals^d} ~~ & \mu \|\vx\|_1 + \frac{1}{2} \sum_{i=1}^m L_1(\vu_i^\top \vx - p_i).
\end{align}
The loss function in the above formulation could be the least squares loss. The related theory and algorithms have been insightfully investigated in the last decade; see \cite{eldar2012compressed} and references therein.

If we only have sign measurements, which could be regarded as an extreme quantization process,  or equivalently $s^+=0$ and $s^-$ being very close to zero in (\ref{saturation-data})\footnote{We assume that the negative measurements are upper bounded by $s^-$.}, we obtain the so-called one-bit compressive sensing (1bit-CS, \cite{boufounos20081} \cite{laska2011trust}). Many 1bit-CS models can be expressed as follows:
\begin{align}\label{1bit-CS}
\min_{\vx \in \reals^d} ~~ & \mu \|\vx\|_{1} + \sum_{i = 1}^m L_2(y_i(\vu_i^\top \vx)) \\
\mbox{s.t.} ~~ & \|\vx\|_{2} = c,\nonumber
\end{align}
where $c$ is a given constant. There are two differences between (\ref{CS}) and (\ref{1bit-CS}). First, a different loss function $L_2$ is used for the inconsistency of the sign. Second, because multiplying a signal by a positive constant does not change the signs of linear measurements, the $\ell_2$ norm constraint ($c=1$ is always assumed) is needed.

In recent works on 1bit-CS (\cite{plan2013robust} \cite{zhang2014efficient} \cite{zhu2015towards}), the nonconvex constraint $\|\vx\|_2 = c$ is relaxed to $\|\vx\|_2 \leq c$ for computational efficiency. But this relaxation makes it inconvenient to use the hinge loss, which is a natural choice for sign inconsistency: minimizing $\sum_{i = 1}^m \max\{0, y_i(\vu_i^\top \vx)\}$ in a norm-ball $\|\vx\|_{2} \leq c$ leads to a trivial and unreasonable solution $\vx = \vzero$. Therefore, in convex 1bit-CS models, $L_2$ usually takes the linear loss $-y_i (\vu_i^\top \vx)$, which could be explained as to pursue high correlation between $y_i$ and $\vu_i^\top \vx$.

\subsection{Mixed One-bit Compressive Sensing}

To consider both analog measurements and one-bit information, we propose the following mixed one-bit compressive sensing (M1bit-CS):
\begin{eqnarray}\label{M1bit-CS}
\min_{\vx \in \reals^d} & & \mu \|\vx\|_{1} + \sum_{i: \Psi_i = 0} L_1(\vu_i^\top \vx - p_i) \nonumber\\
& & ~~~~~~~~ + \lambda \sum_{i: \Psi_i = 1} L_2(y_i(\vu_i^\top \vx - s_i))  \\
\mbox{s.t.} & & \|\vx\|_{2} \leq c. \nonumber
\end{eqnarray}
Instead of putting the $\ell_2$ norm in the constraint, we can also place it in the objective function and come to the following problem,
\begin{align}\label{M1bit-CS-2}
\min_{\vx \in \reals^d} ~ & \mu \|\vx\|_{1} + \frac{\gamma}{2}\|\vx\|^2_{2} \\
 & + \sum_{i: \Psi_i = 0} L_1(\vu_i^\top \vx - p_i)  + \lambda \sum_{i: \Psi_i = 1} L_2(y_i(\vu_i^\top \vx - s_i)), \nonumber
\end{align}
where $\gamma>0$.  This modification is denoted as M1bit-CS with $\ell_2$-norm regularization (\emph{M1bit-CSR}) and the original one (\ref{M1bit-CS}) is M1bit-CS with $\ell_2$-norm constraint (\emph{M1bit-CSC}). When $\|\vx\|_{2}$ is known or can be estimated accurately, one can choose M1bit-CSC, otherwise, M1bit-CSR is applicable.

For the analog part, we use the least squares loss for $L_1$ in this paper. Its robust modifications, such as Huber loss, are also applicable. For the saturation part, directly inheriting the idea of 1bit-CS leads to the linear loss. But here we have also analog measurements that can provide magnitude information, and it makes using the natural hinge loss for sign inconsistency possible. The linear and the hinge loss are covered by the pinball loss defined below:
\begin{equation}\label{pinball}
L_{\tau}(t) = \left\{
\begin{array}{ll}
t, & t \geq 0,\\
-\tau t, & t < 0,
\end{array}
\right.
\end{equation}
where $\tau$ describes the slope in the negative part: when $\tau = 0$, it is the hinge loss and when $\tau = -1$, it reduces to the linear loss. The pinball loss has been applied for both regression (\cite{christmann2007svms} \cite{steinwart2011estimating}) and classification (\cite{huang2014support} \cite{xu2016novel}). For M1bit-CS, it is possible to take any value from $-1$ to $0$ for $\tau$. The best one is problem-dependent and we will investigate its choice in numerical experiments.

In M1bit-CS model, the parameter $\mu$ controls the signal sparsity, and it exists also in regular CS models. This parameter can be selected according to prior-knowledge or by cross-validation. $\lambda$ is a trade-off between the analog and the saturated measurements. Heuristically, we set it as $\lambda = \frac{m}{100 n}$ such that $\lambda$ is smaller when there are more saturated measurements. For M1bit-CSR (\ref{M1bit-CS-2}), there is an additional parameter $\gamma$ which is used to adjust the norm of the reconstructed signal. In 1bit-CS, the requirement on the signal norm is crucial. For M1bit-CS, it becomes less important since there are also analog measurements but still it helps the signal recovery. 

\subsection{Iterative Saturation Detection}\label{sec:ISD}

In the sensing system (\ref{saturation-data}), if the observation is $s^+$ or $s^-$, we know that it is saturated and use it in M1bit-CS as a one-bit measurement. But in some systems, $\vu_i^\top \vx$ itself has a lower/upper bound and then it is interesting and useful to distinguish whether a measurement  reaches the bounds. In this paper, we focus on the lower-saturation and assume that the lower bound is zero. Discussions on upper-saturation and other bounding values are similar.

With prior-knowledge on non-negativeness of the measurements, we have
\[
p_i = s^- ~\Leftrightarrow~ 0 \leq \vu_i^\top \vx \leq s^-.
\]
For example, in a CT scan system, $\vu_i^\top \vx$ is related to the integral of the tissue density along a X-ray. It is certainly non-negative and $\vu_i^\top \vx = 0$ means that the corresponding X-ray hits the detector directly without crossing any tissue. If those observations with $\vu_i^\top \vx = 0$ are found, we can set $\Psi_i = 0, p_i = 0$ and use them as analog measurements in M1bit-CS. They will be helpful, especially for reconstructing the boundary.

The detection criterion is:
\[
\Psi_{i} = \left\{
\begin{array}{ll}
1, & 0 < \vu_i^\top \vx \leq s^-,\\
0, & \vu_i^\top \vx = 0.
\end{array}
\right.
\]
Accordingly, we can use the reconstructed result to update $\Psi$ and then to solve M1bit-CS with the new guess of $\Psi$, iteratively.
Note that incorrectly detecting a saturated measurement as a zero one is more harmful than regarding a zero measurement as a saturated one. Therefore, the initial guess is $\Psi_{i} = 1$ for all $p_{i} = s^-$. For a lower-saturated and non-negative sensing system, the iterative saturation detection (ISD) scheme is described below:

\begin{enumerate}
\item set saturation indicator $\Psi_{i} := 1, \mathrm{if~} p_i \leq s^-$;

\item signal recovery with $\Psi$, denote the result as $\tilde \vx$;

\item calculate the measurements of $\tilde \vx$, i.e., $\tilde q_i := \vu_i^\top \tilde \vx$;

\vskip 0.3cm

\item update
  $
  \Psi_{i} := \left\{
  \begin{array}{ll}
  0, & ~\mathrm{if~} p_{i} > s^-;\\
  0, & ~\mathrm{if~} p_{i} \leq s^- \mathrm{~and~} \tilde q_{i} \leq s^-/10;\\
  1, & ~\mathrm{if~} p_{i} \leq s^- \mathrm{~and~} \tilde q_{i} > s^-/10;
  \end{array}
   \right.
  $

\vskip 0.3cm

\item set $p_i = 0$ for $i: p_i \leq s^- \mathrm{~and~} \Psi_i = 0$;

\item go to 2) until there is no change on $\Psi$.
\end{enumerate}


\section{Alternating Direction Method of Multipliers}
\label{sec:ADMM}
In this paper, we use the least squares loss and the pinball loss for M1bit-CS (\ref{M1bit-CS}) and (\ref{M1bit-CS-2}). Then they are convex problems and many existing methods can be applied. We tried several methods including primal-dual algorithms~\cite{chambolle2011first} and different Alternating Direction Methods of Multipliers (ADMM,~\cite{boyd2011distributed}) implementations~\cite{yan2016self}. In this paper, we introduce an implementation of ADMM, which is efficient in our experiments, but of course is not necessarily optimal for other tasks.

In order to apply ADMM, we reformulate~\eqref{M1bit-CS} by introducing two auxiliary variables $\ve$ and $\vz$, and the equivalent problem is
\begin{eqnarray}\label{model-ADMM}
\min_{\vx, \ve, \vz} & & \mu \|\vx\|_1 + \frac{1}{2} \sum_{i: \Psi_i = 0} (\vu_i^\top \vx - p_i)^2  \nonumber\\
& & ~~ + \lambda \sum_{i: \Psi_i = 1} L_\tau(e_i) + \iota_c(\vz) \\
\mbox{s.t.} & & e_i = y_i(\vu_i^\top \vx - s_i), \forall i: \Psi_i = 1, \nonumber \\
& & \vz = \vx, \nonumber
\end{eqnarray}
where $\iota_c(\vz)$ is an indicator function that returns 0 if $\|\vz\|_2\leq c$ and $+\infty$ otherwise. Note here $\ve$ is $m$-dimensional for consistency but only the $n$ components with $\Psi_i = 1$ have effect.

The corresponding augmented Lagrangian with the dual variables (the Lagrange multipliers: $\valpha$, $\vbeta$) is
\begin{align*}
    &  \cL_A(\vx, \ve, \vz; \valpha, \vbeta) \nonumber \\
  =~&  \mu \|\vx\|_1 + \frac{1}{2} \sum_{i: \Psi_i = 0} (\vu_i^\top \vx - p_i)^2 + \lambda \sum_{i: \Psi_i = 1} L_\tau(e_i) + \iota_c(\vz)  \nonumber \\
    &  + \sum_{i: \Psi_i = 1} \alpha_i \left( e_i - s_i + \vu_i^\top \vx \right) + \frac{\theta_1}{2} \sum_{i: \Psi_i = 1} \left (e_i - s_i + \vu_i^\top \vx \right)^2 \\
    &  + \vbeta^\top \left( \vz - \vx \right) + \frac{\theta_2}{2} \left \|\vz - \vx \right\|_2^2, \nonumber
\end{align*}
where $\theta_1>0$ and $\theta_2>0$.
In every iteration, the ADMM updates $(\ve,\vz)$ and $\vx$ in a Gauss-Seidel style with fixed $(\valpha,\vbeta)$ and updates the dual variable $(\valpha,\vbeta)$.
Updating $(\ve,\vz)$ can be decoupled into updating $\vz$ and $\ve$ independently.
We choose the updating order $(\ve,\vz,\vx,\valpha,\vbeta)$ in our ADMM implementation.
\begin{enumerate}
\item $\ve$-subproblem: we have to solve
\begin{eqnarray*}
\min_\ve & & \lambda \sum_{i: \Psi_i = 1} L_\tau(e_i) + \sum_{i: \Psi_i = 1} \alpha_i \left( e_i - s_i + \vu_i^\top \vx \right) \\
& & + \frac{\theta_1}{2} \sum_{i: \Psi_i = 1} \left (e_i - s_i + \vu_i^\top \vx \right)^2,
\end{eqnarray*}
of which the optimal solution is analytically given by
\[
e_i = \sS_{\tau}\left( s_i - \vu_i^\top \vx - {\alpha_i}/{\theta_1} , {\lambda}/{\theta_1} \right), ~~ \forall i: \Psi_i = 1.
\]
Here $\sS_{\tau}$ is the shrinkage operator for the pinball loss:
\[
\sS_{\tau}(t, \rho) = \left\{
\begin{array}{rl}
t - \rho, & \mbox{if } t \geq \rho, \\
0, & \mbox{if } t \in (\tau \rho,\rho),\\
t - \tau \rho, & \mbox{if } t \leq \tau \rho.
\end{array}
\right.
\]
\item $\vz$-subproblem: we have to solve
\[
\min_\vz~ \textstyle \iota_c(\vz) + \vbeta^\top \vz + \frac{\theta_2}{2} \left \|\vz - \vx \right\|_2^2, \\
\]
of which the analytical solution is
\[
\vz = {\mathcal{P}}_c\left( \vx - {\vbeta}/{\theta_2} \right).
\]
Here ${\mathcal{P}}_c$ is the projection of a vector onto the $\ell_2$ ball $\|\vt\|_2 \leq c$:
\[
{\mathcal{P}}_c(\vt) = \left\{
\begin{array}{rl}
\vt, & \mbox{if }\|\vt\|_2 \leq c, \\
c~{\vt}/{\|\vt\|_2}, &\mbox{if } \|\vt\|_2 > c.
\end{array}
\right.
\]

\item $\vx$-subproblem: we have to solve
\begin{align*}
 \min_{\vx}~ & \mu \|\vx\|_1 + \frac{1}{2} \sum_{i: \Psi_i = 0} (\vu_i^\top \vx - p_i)^2  + (\vU \valpha + \vbeta)^\top \vx \nonumber\\
 & + \frac{\theta_1}{2} \sum_{i: \Psi_i = 1} \left (e_i - s_i + \vu_i^\top \vx \right)^2 + \frac{\theta_2}{2} \left \|\vz - \vx \right\|_2^2.
\end{align*}
It is essentially a quadratic programming plus an $\ell_1$ item:
\begin{align}\label{x_problem}
 \min_{\vx}~ & \textstyle \mu \|\vx\|_1 + \frac{1}{2} \sum_{i: \Psi_i = 0}\limits (\vu_i^\top \vx - p_i)^2   + \frac{\theta_2}{2} \left \| \vx -\vz+{\vbeta\over\theta_2} \right\|_2^2 \nonumber\\
 &\textstyle + \frac{\theta_1}{2} \sum_{i: \Psi_i = 1}\limits \left (e_i - s_i + \vu_i^\top \vx + {\alpha_i\over\theta_1} \right)^2.
\end{align}
It does not have analytical solutions. We approximately solve this problem using FISTA~\cite{beck2009fast}.
\end{enumerate}


The ADMM for M1bit-CSC~\eqref{M1bit-CS} is summarized in Algorithm~\ref{ADMM-M1bit-CS}.
When the $\vx$-subproblem is exactly solved, the convergence of Algorithm~\ref{ADMM-M1bit-CS} follows the classical analysis of ADMM; see, e.g.,~\cite{boyd2011distributed}.
Otherwise, the convergence results for inexact ADMM in~\cite{ng2011inexact} and~\cite{shen2015inexact} can be applied.

\begin{algorithm}[h]
\caption{ADMM for M1bit-CSC (\ref{M1bit-CS})}\label{ADMM-M1bit-CS}
\Indp  Input

  $\vu_i \in \reals^d, \Psi_i \in \{0, 1\}, i = 1, 2, \ldots, m; p_i \in \reals, \forall i: \Psi_i = 0; y_i \in \{-1, +1\}, s_i \in \reals, \forall i: \Psi = 1$; give parameters $\mu, \lambda$ \\

  Initialize

 ~ $\vx, \vz, \vbeta := \vzero_{d}, ~ \ve, \valpha := \vzero_m$, $\theta_1, \theta_2 > 0$

\Repeat{stopping criteria is satisfied}
{

  $e_i := \sS_{\tau}\left( s_i - \vu_i^\top \vx - {\alpha_i}/{\theta_1} , {\lambda}/{\theta_1} \right), \forall i: \Psi_i = 1$

  $\vz := {\mathcal{P}}_c\left( \vx - {\vbeta}/{\theta_2} \right)$

  update $\vx$ by solving~\eqref{x_problem}

  $\valpha := \valpha + \theta_1 (\ve - \vs + \vU^\top \vx)$

  $\vbeta := \vbeta + \theta_2 (\vz - \vx)$
}
\end{algorithm}


We can establish ADMM in the same way for M1bit-CSR~\eqref{M1bit-CS-2}, and the algorithm is similar to Algorithm~\ref{ADMM-M1bit-CS} except that the update of $\vz$ becomes solving the following problem:
\[
\min_\vz~ \frac{\gamma}{2} \|\vz\|^2_2 + \vbeta^\top \vz + \frac{\theta_2}{2} \left \|\vz - \vx \right\|_2^2, \\
\]
of which the optimal solution is analytically given by
\begin{equation}\label{RCSS_m}
\vz = \frac{\theta_2 \vx - \vbeta}{\theta_2 + \gamma}.
\end{equation}

\section{Experimental Validation}\label{sec:numerical}

In this section, we evaluate the proposed algorithms on one-dimensional sparse signal recovery problems. The parameters will be discussed numerically, and then the algorithms will be compared with two existing methods: i) simply dropping those saturated measurements and using the remaining measurements to recover the sparse signal by lasso;  ii) {Robust Dequantized Compressive Sensing (RDCS,~\cite{liu2014robust}), which is designed for both saturation and quantization error. When there is no quantization error, RDCS takes the following formulation:
\begin{align}\label{RDCS}
\min_{\vx \in \reals^d} ~~ & \mu \|\vx\|_1 + \frac{1}{2} \sum_{i: \Psi_i = 0} (\vu_i^\top\vx- p_i)^2 \nonumber\\
\mbox{s.t.} ~~ & y_i(\vu_i^\top \vx - s_i) \geq 0, ~ \forall i: \Psi_i = 1.
\end{align}
It assumes that there is no noise in the saturated measurements. Thus model~\eqref{RDCS} is not suitable when there is noise in measurements before quantization.
In fact, model~\eqref{RDCS} is equivalent to model~\eqref{M1bit-CS} with $\tau=0$, $\lambda=+\infty$, and $c=+\infty$. The ADMM algorithms for lasso and RDCS can be found in~\cite{boyd2011distributed} and~\cite{liu2014robust}, respectively. All the experiments are done with Matlab 2014b on Windows 7 with Core i5-3.10 GHz and 8.0 GB RAM.

Let us consider an experiment with $d = 1000$. Denote the true signal as $\bar \vx$, of which the non-zero components are generated firstly following the standard Gaussian distribution and then normalized such that $\|\bar \vx\|_{2} = 1$.
The sensing vectors are drawn from the standard Gaussian distribution. There is Gaussian noise in the measurements. The noise level is measured by the ratio of the summed squared magnitude of noise-free measurements to that of noise and denoted by $s_n$.
For the saturation part, we choose the saturation levels such that there are $n/2$ upper saturated measurements and $n/2$ lower saturated ones.
Suppose that the reconstructed signal is $\tilde \vx$, its quality is measured by the signal-to-noise ratio (SNR) in dB:
\[
\mathrm{SNR}(\bar \vx, \tilde \vx) = 10 \log_{10} \left( {\|\bar \vx\|^2_{2}}/{\|\bar \vx - \tilde \vx\|^2_{2}} \right).
\]


\subsection{Parameters Selection}
The parameter $\tau$ in the pinball loss is important to characterize the M1bit-CS model. As discussed previously, $\tau = -1$ is for 1bit-CS, $\tau = 0$ is for classification problems. To numerically investigate the performance of different $\tau$ values, the following situations are considered:
\begin{enumerate}
\item signal sparsity $K = 100$, number of measurements $m = 500$, number of saturated observations $n = 100$, noise level $s_n = 20$;
\item $K, m, s_n$ are as the same as 1), but $n = 50$;
\item $K, m, s_n$ are as the same as 1), but $n = 200$;
\item $K, m, n$ are as the same as 1), but $s_n = 10$.
\end{enumerate}
We then apply Algorithm \ref{ADMM-M1bit-CS} to recover the signal and report the average SNR over 100 trials in Fig.\ref{fig-tau}. From the result, one can observe that a small and negative $\tau$ performs well. The trend is that when the saturation ratio $m/n$ is large, the absolute value of the best $\tau$ is also large, which coincides with the experiments reported in 1bit-CS literatures: $\tau = -1$ is better than $\tau = 0$ in 1bit-CS where $m = n$. In the rest of this paper, we always use $\tau = -\frac{n}{5m}$. Additional tuning can improve the performance but requires more computational burden.

\begin{figure}[htbp]
  \centering
   \psfrag{snr}[c]{\footnotesize SNR}
   \psfrag{tau}[c]{\footnotesize $\tau$}
   \includegraphics[width=0.6\linewidth]{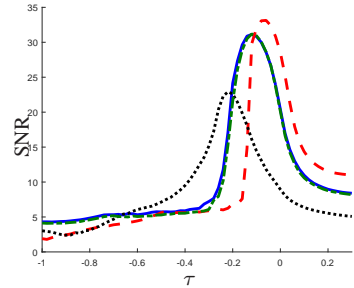}
  \caption{Reconstructed SNRs for different $\tau$ values. CASE 1): blue solid curve; CASE 2): red dashed curve; CASE 3): black dotted curve; CASE 4): green dot-dashed curve.}\label{fig-tau}
\end{figure}

For M1bit-CSR (\ref{M1bit-CS-2}), there is an additional parameter $\gamma$ which is used to adjust the norm of the reconstructed signal. In 1bit-CS, the requirement on the signal norm is crucial. For M1bit-CS, it becomes less important since there are also analog measurements but still it helps the signal recovery. We in the following consider different $\gamma$ values for $d = 1000, K = 100, m = 500, n = 400$, and $s_n = 10$. The average SNR over 100 trials for different $\gamma$ values are displayed by the blue solid curve in Fig.\ref{fig-gamma}, which also shows the norm of the reconstructed signals by the red dashed curve. As expected, the highest SNR is obtained when $\|\tilde \vx\|_2$ is near one. We in this paper use $\gamma = 10^{-4}$. Note that when there are fewer  saturation data, the influence of the norm constraint is weaken. In that case, $\|\vx\|_2$ could be simply regarded as a regularization term.

\begin{figure}[htbp]
  \centering
   \psfrag{SNR}[c]{\footnotesize SNR}
   \psfrag{gamma}[c]{\footnotesize $\log_{10}(\gamma)$}
   \psfrag{norm}[c]{\footnotesize $\|\tilde \vx\|_2$}
   \includegraphics[width=0.6\linewidth]{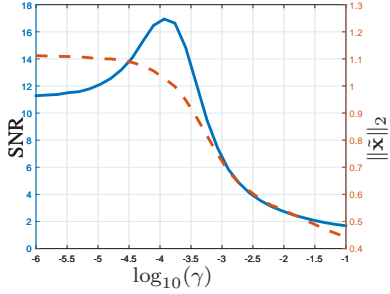}
  \caption{Reconstruction performance for different $\gamma$ values. SNR: blue solid curve; The $\ell_2$ norm of the reconstructed signal: red dashed curve.}\label{fig-gamma}
\end{figure}

%
%
%

\subsection{Synthetic Experiments}
In this subsection, we compare M1bit-CS with lasso and RDCS on synthetic data. The sparsity parameter $\lambda$ is tuned based on lasso, and the same value is then applied to other methods. We first set $d = 1000$, $K = 300$, $m = 500$, $s_n = 10$ and consider the recovery performance with different saturation ratios ${n}/{m}$ varying from $0\%$ to $40\%$.
The average results over $100$ trials are displayed in Fig.\ref{fig-r:a}.
When the saturation ratio is low, there are plenty of analog measurements, then lasso can reconstruct the signal well.
When $n/m$ increases, lasso's recovery quality dramatically decreases.
Saturated measurements are helpful to improve the recovery performance, which is the reason why RDCS, M1bit-CSC, and M1bit-CSR have significantly better results than lasso.
Among the three algorithms, M1bit-CSC performs the best. Its advantage over RDCS is due to the facts that noise on saturated measurements are suppressed and the signal's norm information is useful for recovery.
If the $\ell_2$ norm of the true signal is unknown or cannot be well estimated, then we use M1bit-CSR, whose performance is slightly worse than M1bit-CSC.
Similar performance could be observed in Fig.\ref{fig-r:b}, which is the average recovery results over 100 trails for different sparse levels with a fixed saturation ratio $n/m = 20\%$.

\begin{figure}[htbp]
  \centering
  \subfigure[]{
    \psfrag{snr}[c]{\footnotesize SNR (dB)}
    \psfrag{ratio}[c]{\footnotesize saturation ratio $n/m$}
    \psfrag{data1}[l]{\tiny lasso}
    \psfrag{data2}[l]{\tiny RDCS}
    \psfrag{data3}[l]{\tiny M1bit-CSC}
    \psfrag{data4}[l]{\tiny M1bit-CSR}
    \label{fig-r:a} 
    \includegraphics[width=0.47\linewidth]{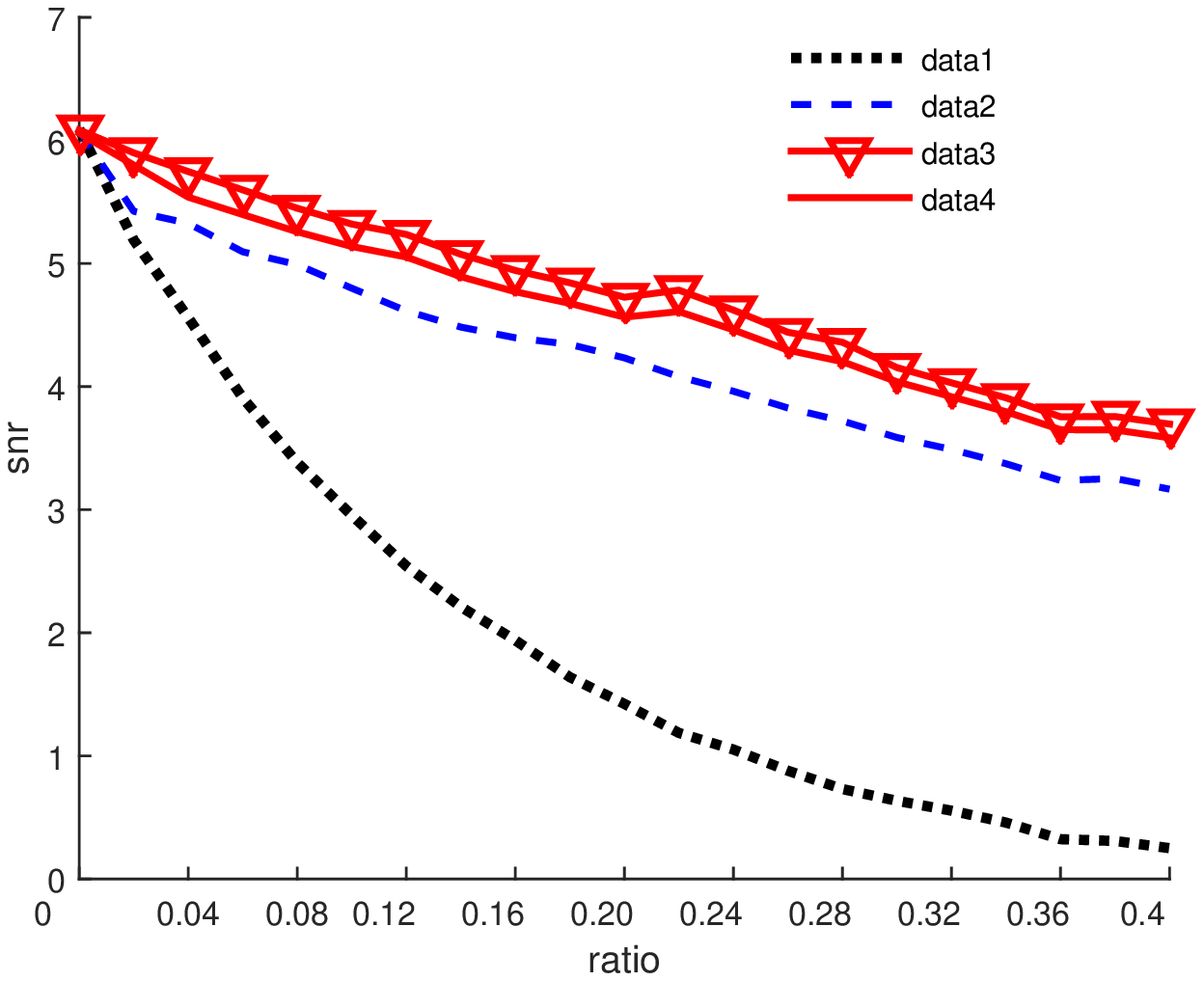}}
  \subfigure[]{
    \psfrag{snr}[c]{\footnotesize SNR (dB)}
    \psfrag{K}[c]{\footnotesize number of non-zero components}
    \psfrag{data1}[l]{\tiny lasso}
    \psfrag{data2}[l]{\tiny RDCS}
    \psfrag{data3}[l]{\tiny M1bit-CSC}
    \psfrag{data4}[l]{\tiny M1bit-CSR}
    \label{fig-r:b} 
    \includegraphics[width=0.47\linewidth]{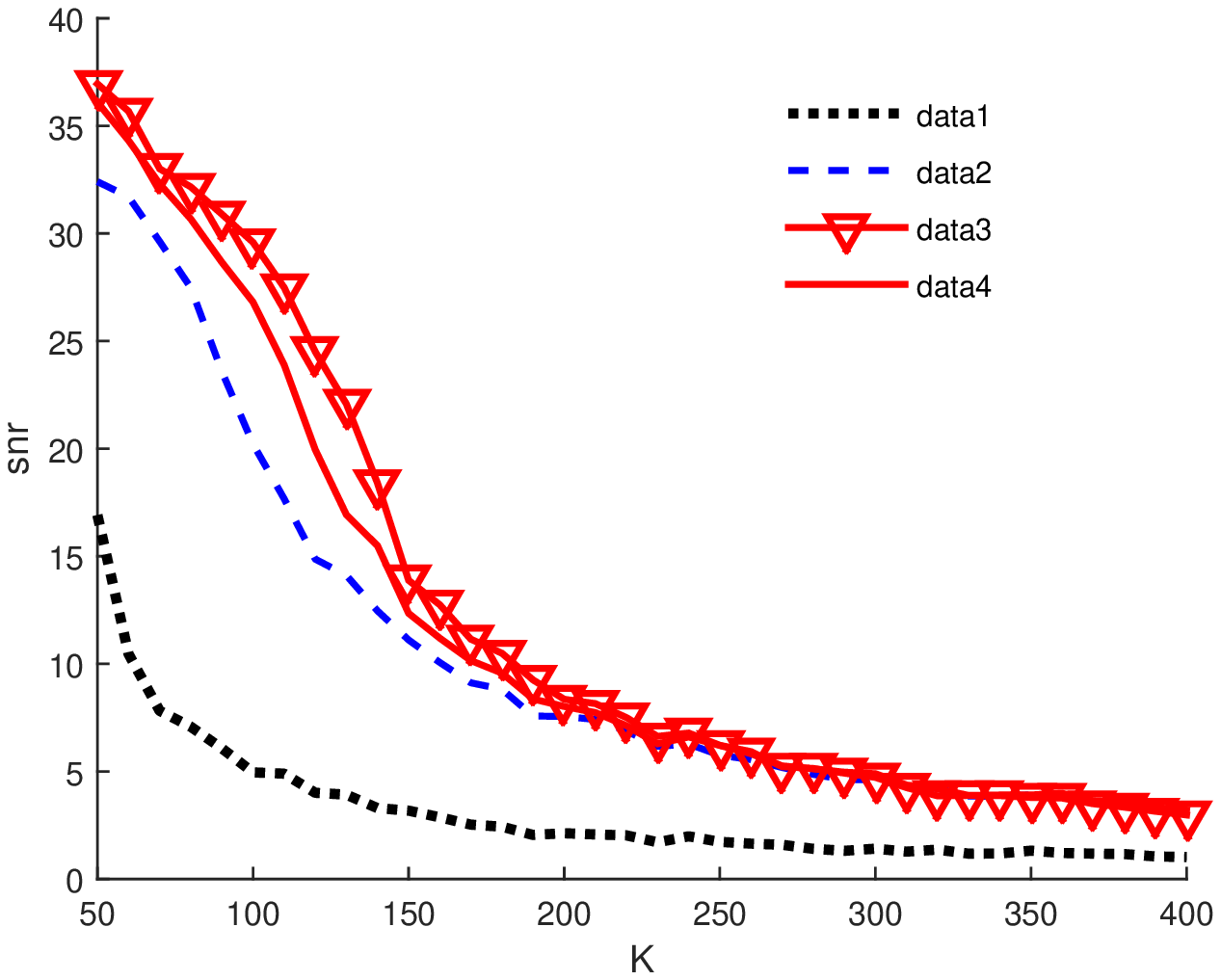}}
  \caption{Recovery performance of lasso (black dotted), RDCS (blue dashed), M1bit-CSC (red solid line with triangle), and M1bit-CSR (red solid) for (a) different saturation ratios and (b) different numbers of non-zero components.}\label{fig-r}
\end{figure}

We then keep the signal sparsity as $K = 300$, the saturation ratio as ${n}/{m} = 20\%$, and vary the number of measurements $m$ from $350$ to $2000$.
The average SNRs over 100 trials are reported in Fig.\ref{fig-m:a}.
We can observe the effect of the proposed algorithms in mining knowledge from saturated measurements.
For example, M1bit-CSC with $800$ measurements (640 analog measurements and 160 saturated ones) has a similar SNR as lasso with 1000 measurements (800 analog measurements are used and the left 200 ones are dropped).
It means that saturated measurements have almost the same power as analog ones in recovering the sparse signal in that case and our algorithms are able to explore all the power with a longer computational time. Generally, the computational time of M1bit-CS is about 10 times of that of lasso, as reported in Fig.\ref{fig-m:b}.
We think that when there are saturated measurements and if the computational time requirement is not critical, it is worthy to use M1bit-CS to improve the recovery accuracy.
\begin{figure}[htbp]
  \centering
  \subfigure[]{
    \psfrag{snr}[c]{\footnotesize SNR (dB)}
    \psfrag{m}[c]{\footnotesize number of measurements}
    \psfrag{data1}[l]{\tiny lasso}
    \psfrag{data2}[l]{\tiny RDCS}
    \psfrag{data3}[l]{\tiny M1bit-CSC}
    \psfrag{data4}[l]{\tiny M1bit-CSR}
    \label{fig-m:a} 
    \includegraphics[width=0.47\linewidth]{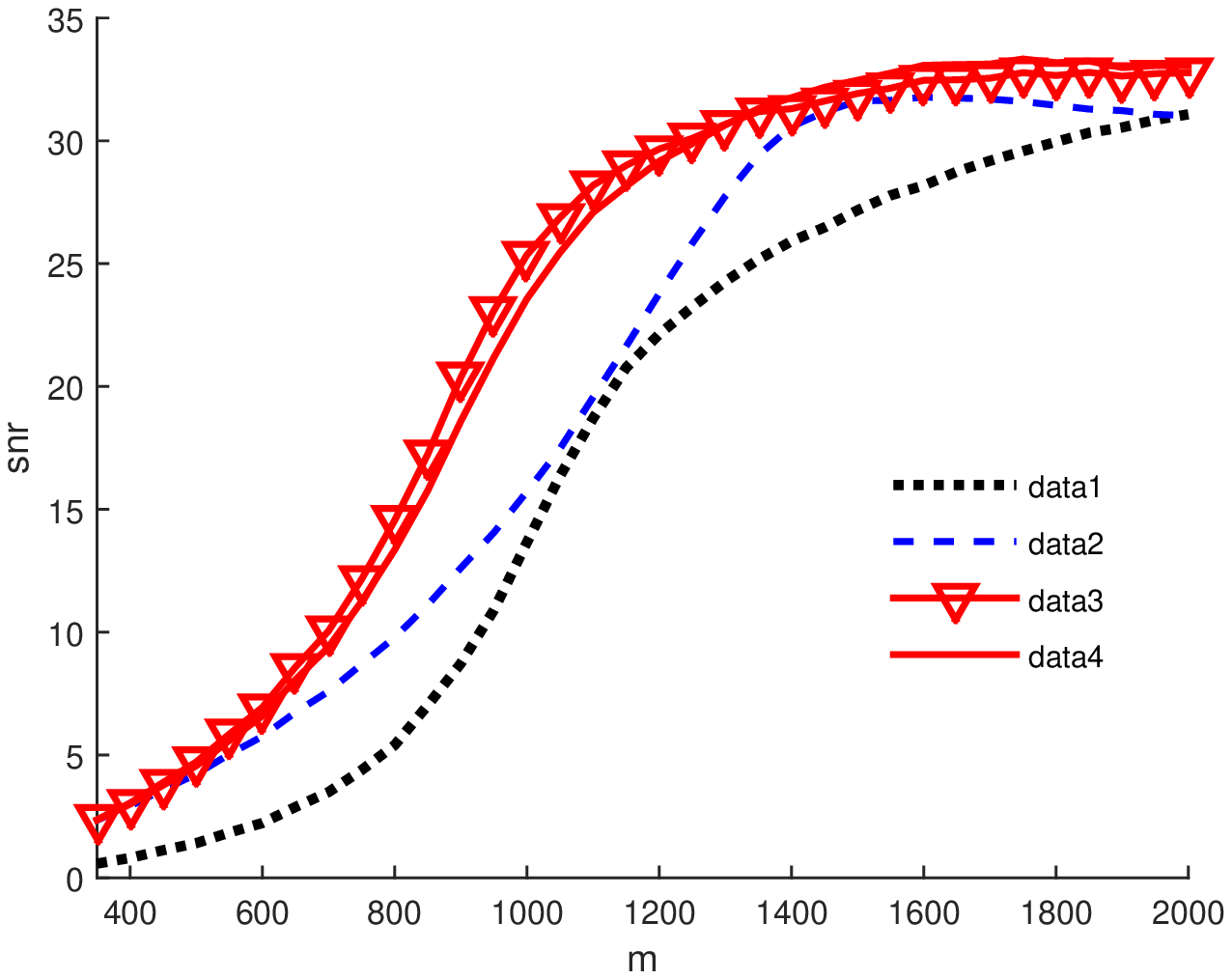}}
  \subfigure[]{
    \psfrag{time}[c]{\footnotesize time (s)}
    \psfrag{m}[c]{\footnotesize number of measurements}
    \psfrag{data1}[l]{\tiny lasso}
    \psfrag{data2}[l]{\tiny RDCS}
    \psfrag{data3}[l]{\tiny M1bit-CSC}
    \psfrag{data4}[l]{\tiny M1bit-CSR}
    \label{fig-m:b} 
    \includegraphics[width=0.47\linewidth]{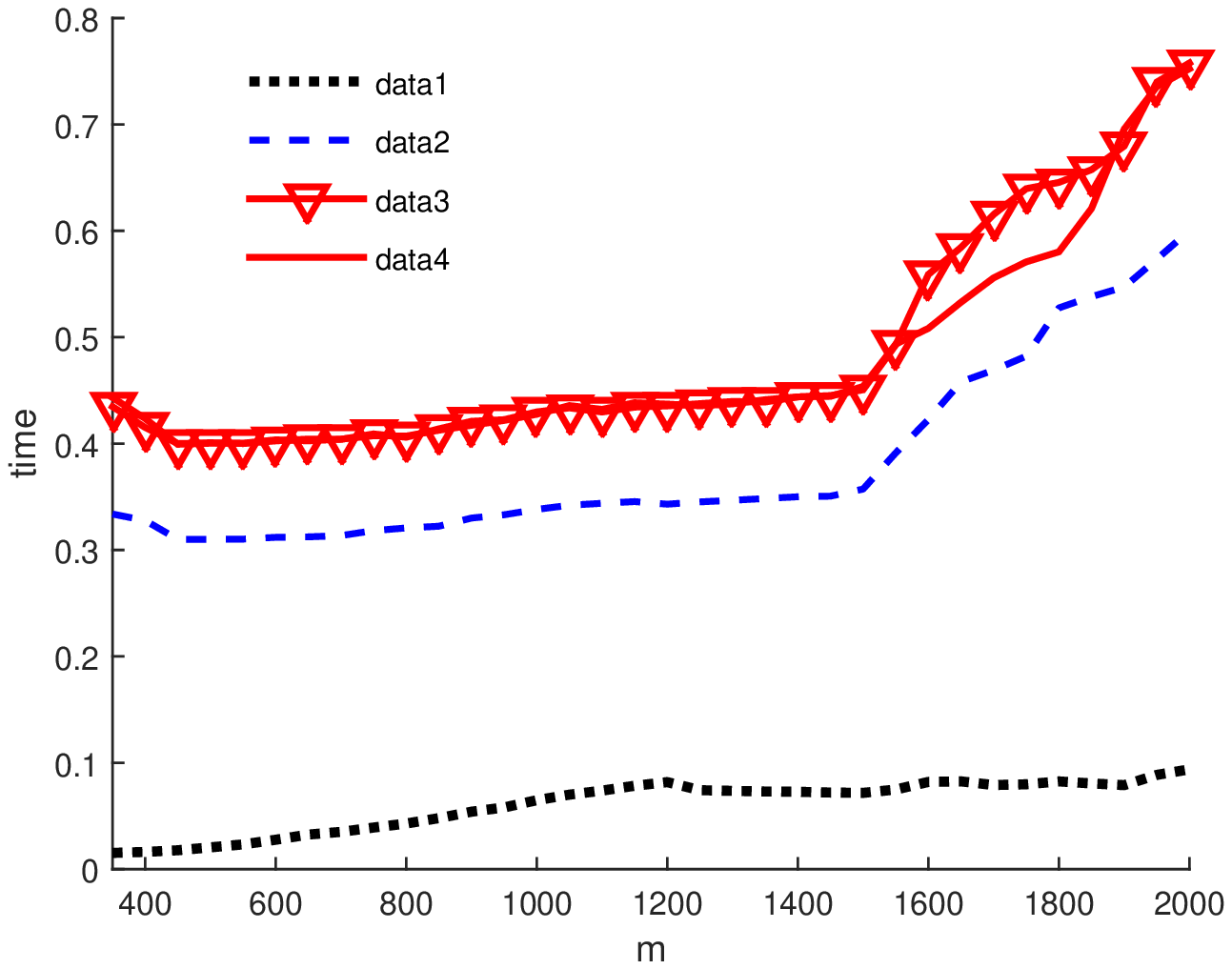}}
  \caption{Recovery performance of lasso (black dotted), RDCS (blue dashed), M1bit-CSC (red solid line with triangle), and M1bit-CSR (red solid) for different numbers of measurements. (a) SNR; (b) computational time.}\label{fig-m}
\end{figure}

\section{Overexposure correction for CT Reconstruction}
\label{sec:CT}

\subsection{Problem Formulation}\label{sec:CT-problem}
Now we are at the stage to apply the proposed M1bit-CS to CT reconstruction to correct overexposure. In a CT system, the object density image can be represented in its discrete form as a vector $\vx$, and the acquired projection data as $\vq$. When there is no saturation, the X-ray transform of the object $\vx$ can be written as
\begin{equation}
\vq = \vU^\top \vx,
\end{equation}
where $\vU$ is the system matrix.
Suppose that the intensity of one X-ray before entering the object is $I_{0}$ and the intensity departing the object is $I_{1}$, then Lambert-Beer's law tells us:
\[
I_{1}=I_{0}e^{-q}.
\]
Due to the physical limits of the detector or the analog-to-digital converter, there is a maximal intensity $I_{\max}$ and any intensity exceeding this threshold cannot be distinguished.
In other words, we cannot observe the analog value when
\[
q < \log(I_{0}/I_{\max}).
\]
In practice, the maximal and minimal detectable value of the detector can be adjusted at each projection angle for exposure control but the measurable dynamic range of the detector is fixed and denoted by $\kappa = \log (I_{\beta,\max}/ I_{\beta,\min})$, where $I_{\beta,\max}$ and $I_{\beta,\min}$ are the maximal and minimal intensities that are measured by the detector at projection angle $\beta$. At each projection, $I_{\beta,\min}$ is determined by the object, and $I_{\beta,\max}$ is adjusted correspondingly afterwards by the fixed $\kappa$ as $I_{\beta,\max} = I_{\beta,\min} e^\kappa$. Then we can choose a dynamic threshold $s_\beta =\log(I_0/I_{\beta,\max})=\log(I_0/I_{\beta,\min})-\kappa = p_{\beta,\max} - \kappa$, where $p_{\beta,\max}$ is the maximum projection value after the minus log transform at projection $\beta$. Thus, the observation $\vp$ is a truncation of $\vq$, i.e., $p_i = \max\{s_i, q_i\}$, where $s_i$ takes the value of $s_\beta$ with $\beta$ being the angle containing the $i$-th projection. Since the observation stands for the integral of the tissue intensity, it should be non-negative. In fact, there are amount of X-rays that hit the detector directly, i.e., the corresponding measurements is zero. Thus, when a measurement get saturated, people usually set its value as zero. Therefore, the observations in CT reconstruction are
\[
p_{i} =
\left\{
\begin{array}{ll}
q_{i}, & \mathrm{if~~} q_{i} > s_{i}, \\
0,     & \mathrm{if~~} q_{i} \leq s_{i}.
\end{array}
\right.
\]

When overexposure happens, traditional CT reconstruction methods are not suitable, as shown by Fig.\ref{fig-phantom:b}. To correct overexposure, the researchers typically considered extrapolation to compensate the missing information; see e.g., \cite{Ohnesorge00:ECF} \cite{Maier12:OEF} \cite{Xia14TCA}. Among them, the method given by \cite{Hsieh04:ANR} that assumes the missing data as line integrals of a partial water cylinder is efficient. It is based on FBP algorithm and in the following, we refer this method as FBP-WCE (FBP with water cylinder extrapolation). Its reconstruction result is displayed in Fig.\ref{fig-phantom-M1bit:a}. One can see significant improvement from the standard FBP method, of which the result is given by Fig.\ref{fig-phantom:b}.

\subsection{Knee Phantom}

The overexposure phenomenon is a kind of one-sided saturation. If $\Psi$ is known, we can directly apply M1bit-CSR (\ref{M1bit-CS-2}) to do overexposure correction (we do not know the $\ell_2$ norm of the image and M1bit-CSC is not suitable). Note that the sparsity for CT reconstruction is on the gradient domain, i.e., total variation (TV) is used as the regularization term; see \cite{yu2010soft} \cite{ritschl2011improved} \cite{chen2013limited} for TV regularization CT reconstruction. When applying M1bit-CS on CT reconstruction, we do not use FISTA but a TV minimization algorithm to solve the $\vx$-subproblem (\ref{x_problem}). The reconstruction image is shown in Fig.\ref{fig-phantom-M1bit:b}, which preliminarily shows that acquiring information from saturated projections largely helps CT reconstruction.

\begin{figure}[htbp]
  \centering
  \subfigure[]{\label{fig-phantom-M1bit:a}
    \includegraphics[width=0.4\linewidth]{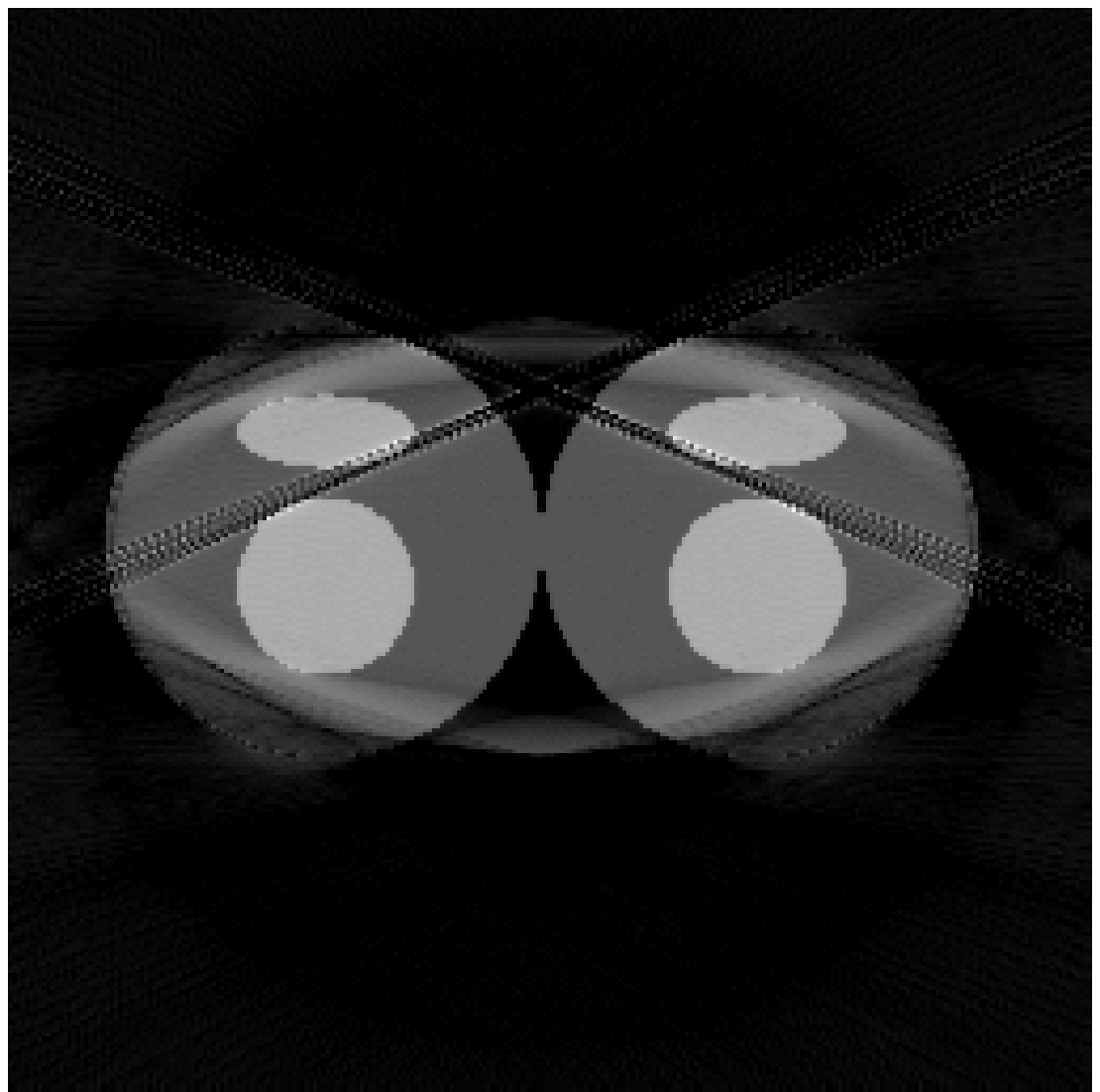}} \quad
  \subfigure[]{\label{fig-phantom-M1bit:b}
    \includegraphics[width=0.4\linewidth]{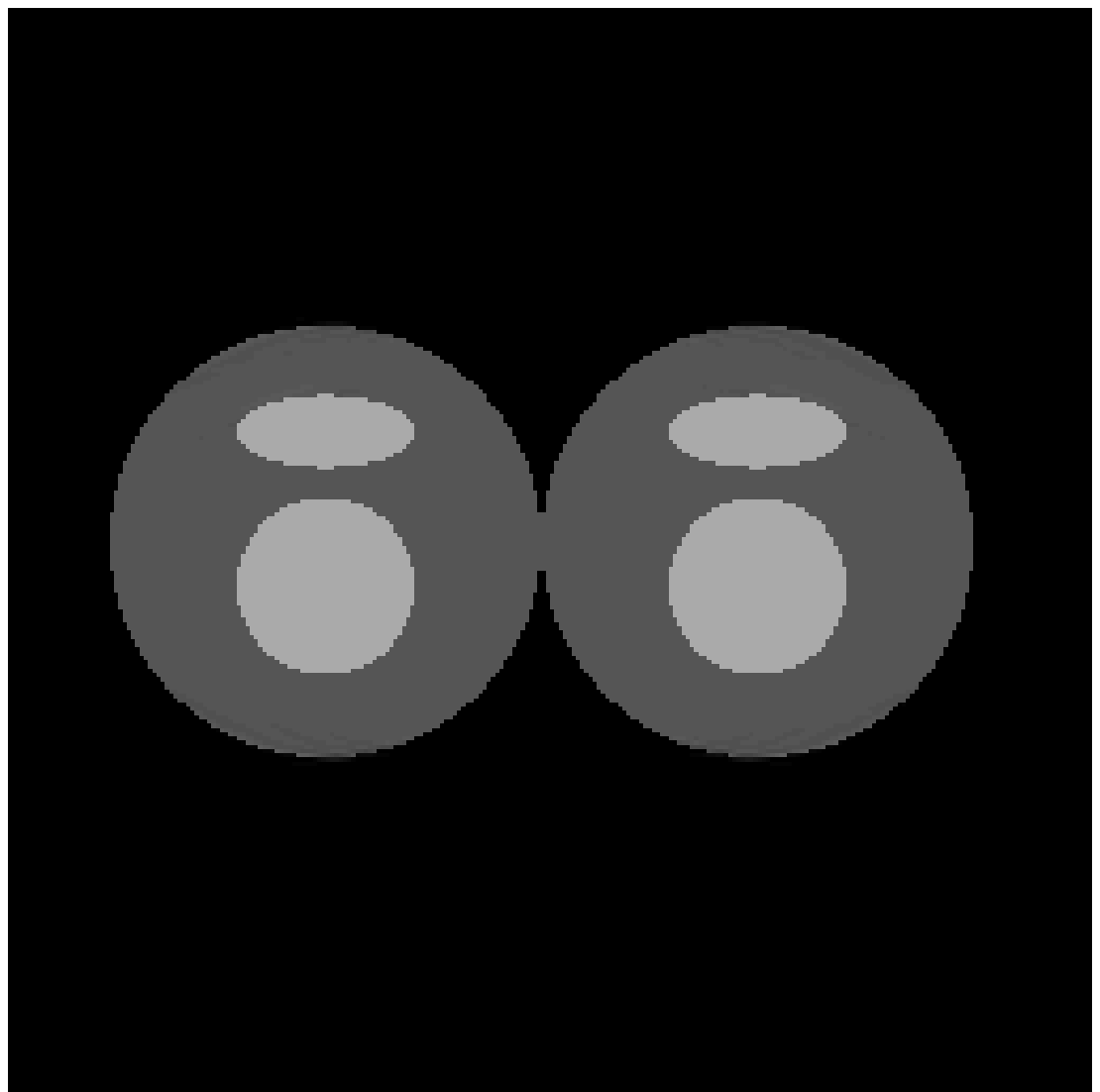}}
  \caption{Reconstructed results from saturated projection Fig.\ref{fig-phantom-projection:b} (a) FBP-WCE; (b) M1bit-CSR.}\label{fig-phantom-M1bit}
\end{figure}

In practice, when $p_i = 0$, we do not know whether this projection ray is saturated or does not hit the object. Thus, the ISD scheme proposed in Section \ref{sec:ISD} should be used together with M1bit-CSR  (abbreviated as M1bit-CSR-ISD). Again consider the projections shown in Fig.\ref{fig-phantom-projection:b}. Fig.\ref{fig-phantom-saturation_matrix:b} displays the detected $\Psi$. The difference between the ideal and the detected $\Psi$ is shown in Fig.\ref{fig-phantom-saturation_matrix:c}, where the false detections (zero measurements that are detected as saturated ones) are marked in blue and missing detection (saturated measurements that are regarded as zero ones) are in green.

\begin{figure}[htbp]
  \centering
  \subfigure[]{\label{fig-phantom-saturation_matrix:a}
    \includegraphics[width=0.75\linewidth]{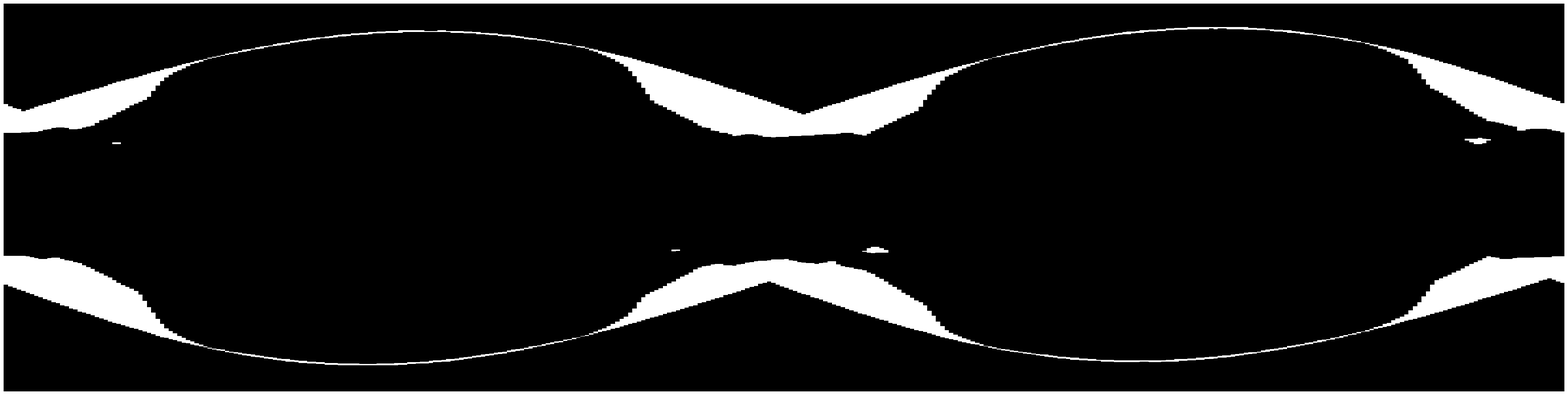}} \quad
  \subfigure[]{\label{fig-phantom-saturation_matrix:b}
    \includegraphics[width=0.75\linewidth]{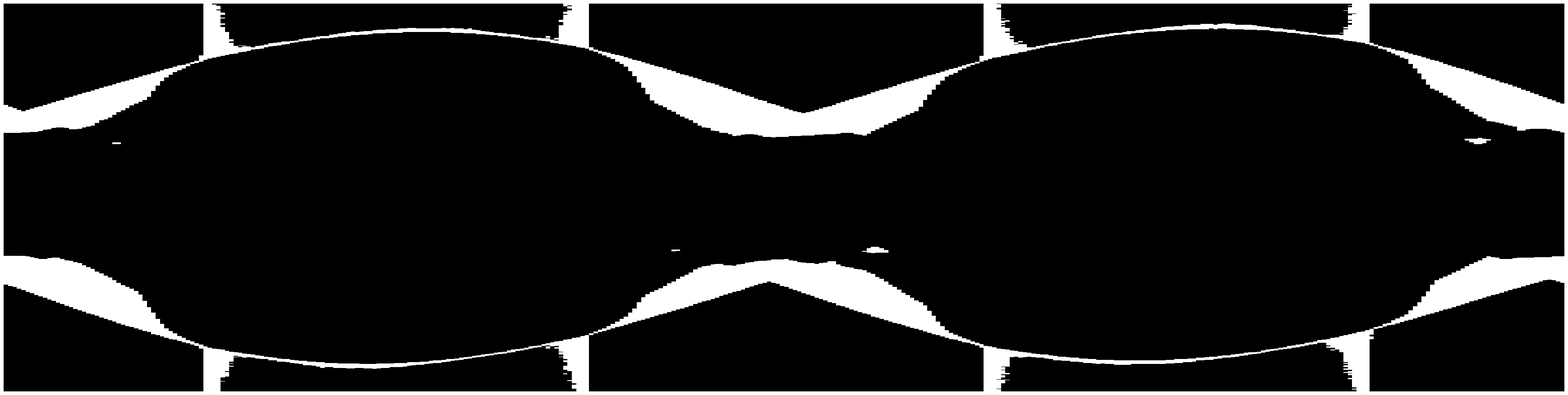}}
  \subfigure[]{\label{fig-phantom-saturation_matrix:c}
    \includegraphics[width=0.75\linewidth]{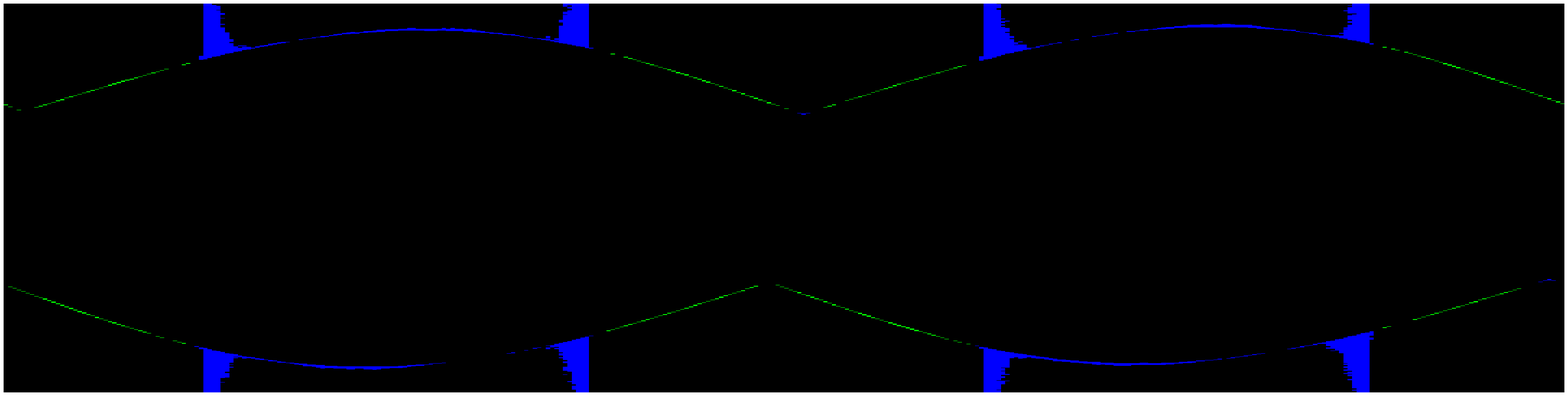}}
  \caption{Saturation indicator $\Psi$ for saturated data shown in Fig.\ref{fig-phantom-projection:b}: (a) the true $\Psi$; (b) indicator detected by M1bit-CSR-ISD; (c) difference between the true and the detected saturation indicator (blue: false detection; green: missing detection).}\label{fig-phantom-saturation_matrix}
\end{figure}

The final reconstruction image is given by Fig.\ref{fig-phantom-M1bit-detection}, which shows that a small number of incorrect detections can be tolerated by M1bit-CSR (\ref{M1bit-CS-2}) and the reconstruction result is still satisfactory. As plotted in Fig.\ref{fig-phantom-M1bit-detection:b}, root of mean square error (RMSE, in HU) is decreasing with the increasing of ISD iterations. Since ISD requires to solving M1bit-CSR multiple times, the detection process is time-consuming. Utilizing prior-knowledge may reduce the detection time, which could be an interesting study in the future.

\begin{figure}[htbp]
  \centering
  \subfigure[]{\label{fig-phantom-M1bit-detection:a}
    \includegraphics[width=0.4\linewidth]{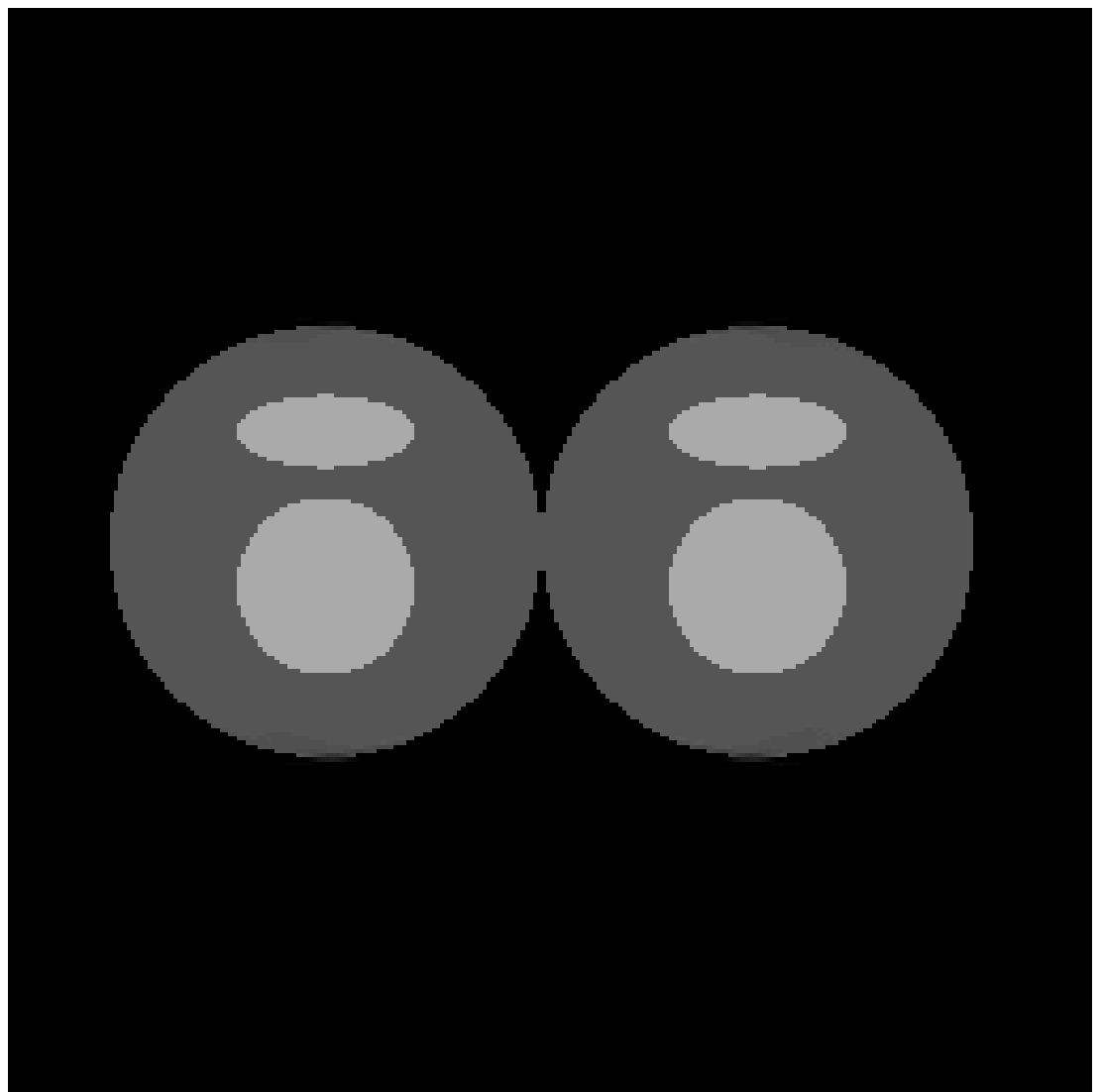}} \quad
  \subfigure[]{\label{fig-phantom-M1bit-detection:b}
    \psfrag{RMSE (HU)}[c]{\footnotesize RMSE (HU)}
    \psfrag{ISD iteration}[c]{\footnotesize ISD iteration}
    \includegraphics[width=0.44\linewidth]{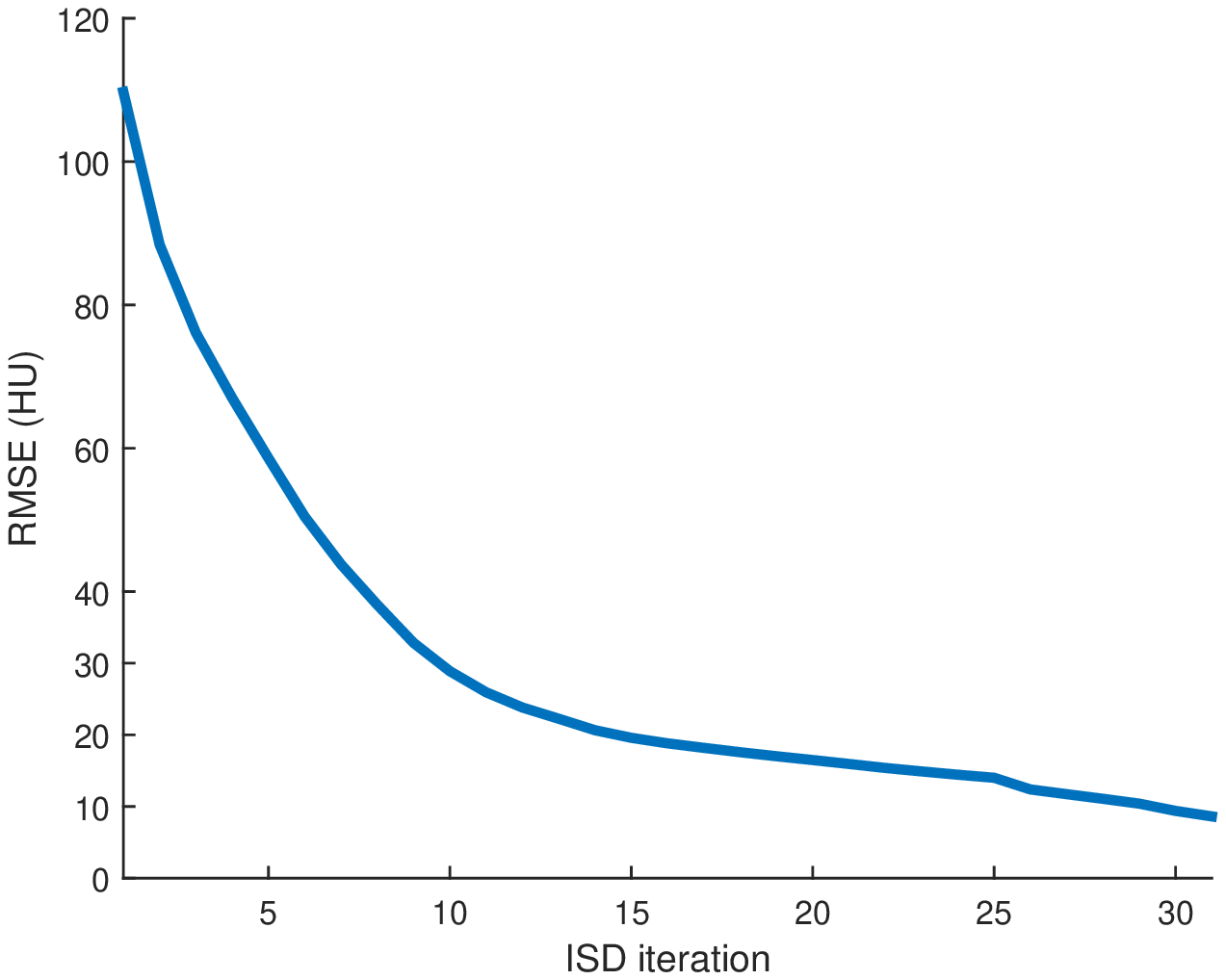}}
  \caption{(a) Reconstructed result of M1bit-CS-ISD from saturated projections shown in Fig.\ref{fig-phantom-projection:b}; (b) RMSE (HU) in different ISD iterations.}\label{fig-phantom-M1bit-detection}
\end{figure}

\subsection{Simulated Clinical Data}

Finally, we apply the proposed method on a clinical head dataset. The data are acquired with a Siemens Artis zee angiographic C-arm system (Siemens Healthcare GmbH, Forchheim, Germany). In this experiment, we choose one slice of a 3D clinical head dataset as the ground truth image (Fig.\ref{fig-real-result-15:0}) and reproject it to simulate the acquired sinogram data in the fan-beam system with the following trajectory parameters: the source-to-isocenter distance is 750 mm and isocenter-to-detector distance is 450 mm. The angular step is 1 degree and the total scan range is 360 degrees. The equal-spaced detector length is 620 mm with the pixel length $1$ mm.


The full projections are shown in Fig.\ref{fig-real-projection:a}. When the detection range $\kappa$ is limited, there could be saturation for the projection. In Fig.\ref{fig-real-projection:b} and Fig.\ref{fig-real-projection:c}, observations for $\kappa = 0.6 p_{\max}$ and $\kappa = 0.4 p_{\max}$ are displayed, respectively. Our task is to recover the image from the saturated projections via M1bit-CS-ISD. The results are compared with FBP and SART, two standard CT reconduction frameworks. For FBP, we apply the modification given by \cite{Hsieh04:ANR} that utilizes water cylinder extrapolation to remedy missing projections caused by truncation or overexposure. For SART, one can remove those saturated projections when they are found, for which the ISD can be used as well. We denote this method as SART-ISD, of which the detection scheme is as the same as M1bit-CSR-ISD but SART is used as the reconstruction method.



\begin{figure}[htbp]
  \centering
  \subfigure[]{\label{fig-real-result-15:0}
    \includegraphics[width=0.45\linewidth]{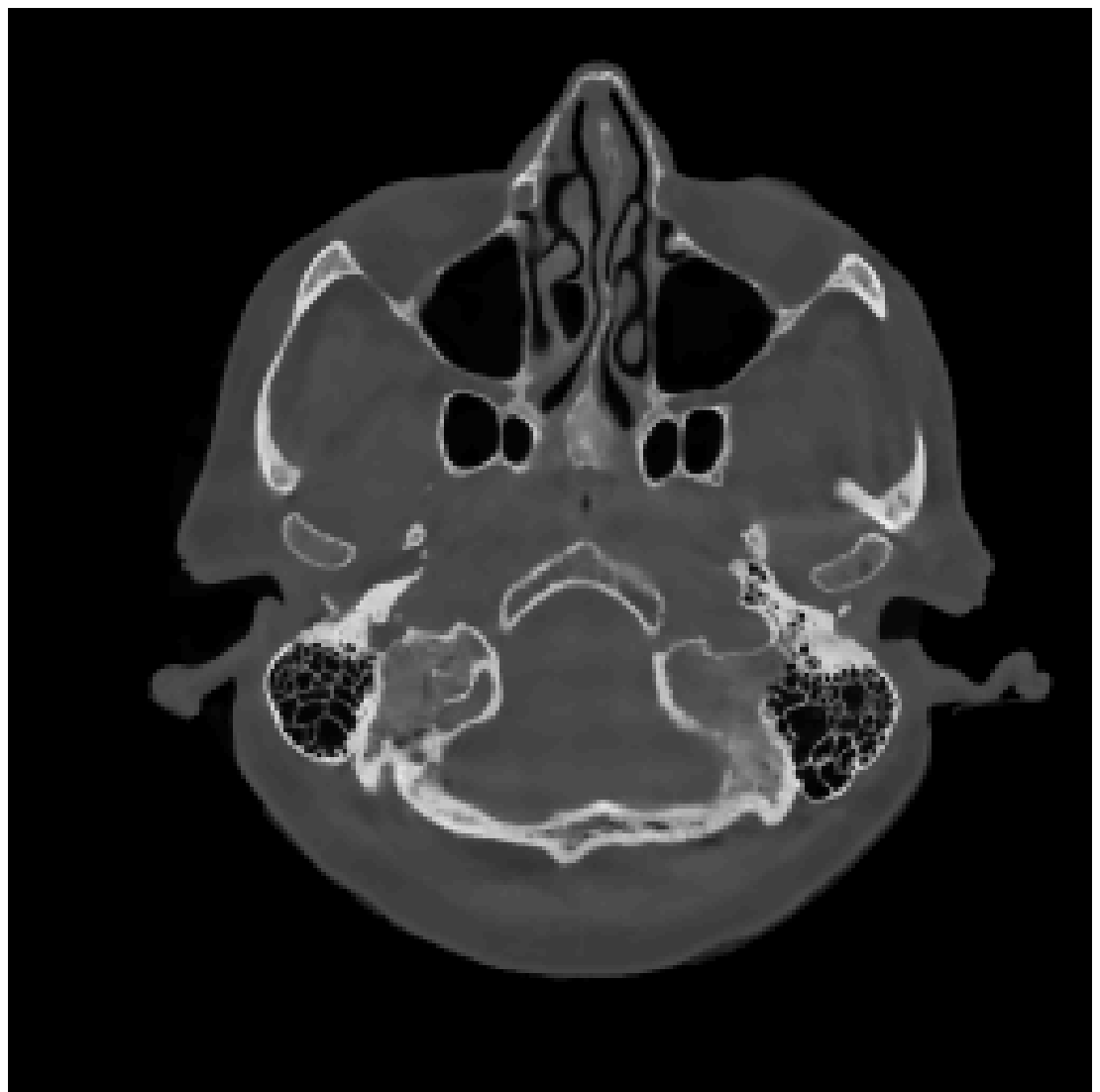}}\quad
  \subfigure[]{\label{fig-real-result-15:a}
    \includegraphics[width=0.45\linewidth]{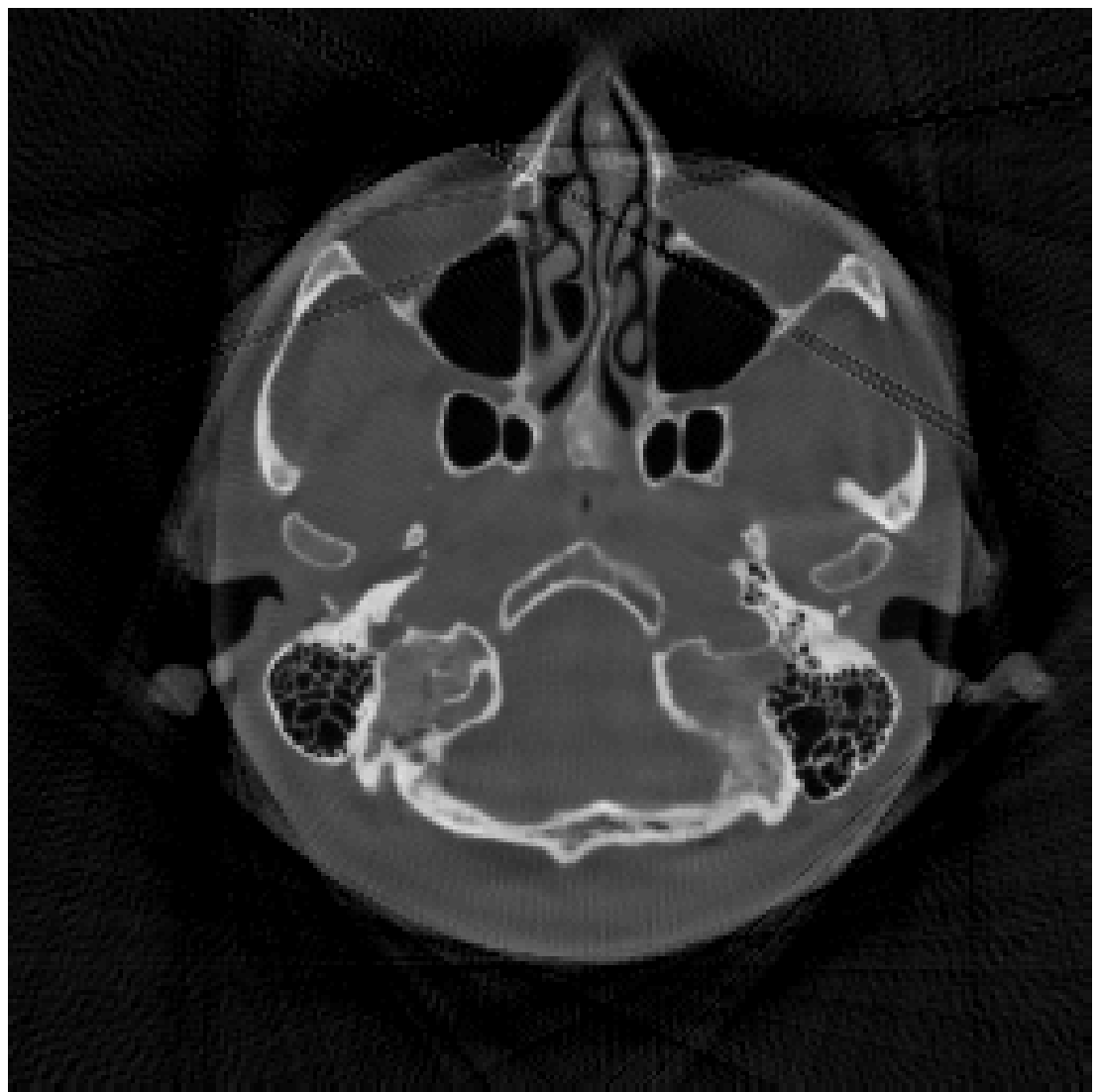}} \\
  \subfigure[]{\label{fig-real-result-15:b}
    \includegraphics[width=0.45\linewidth]{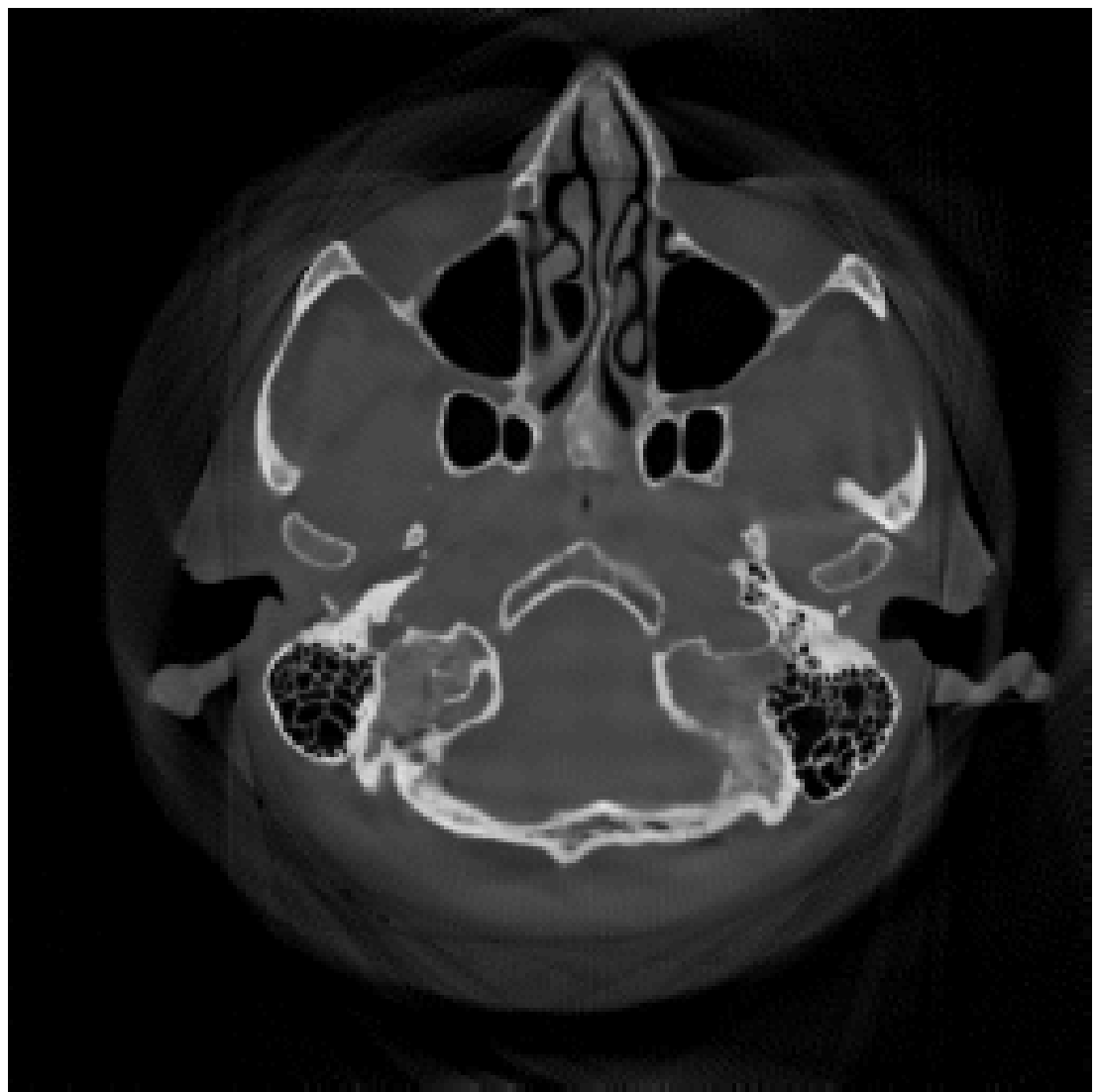}}\quad
  \subfigure[]{\label{fig-real-result-15:c}
    \includegraphics[width=0.45\linewidth]{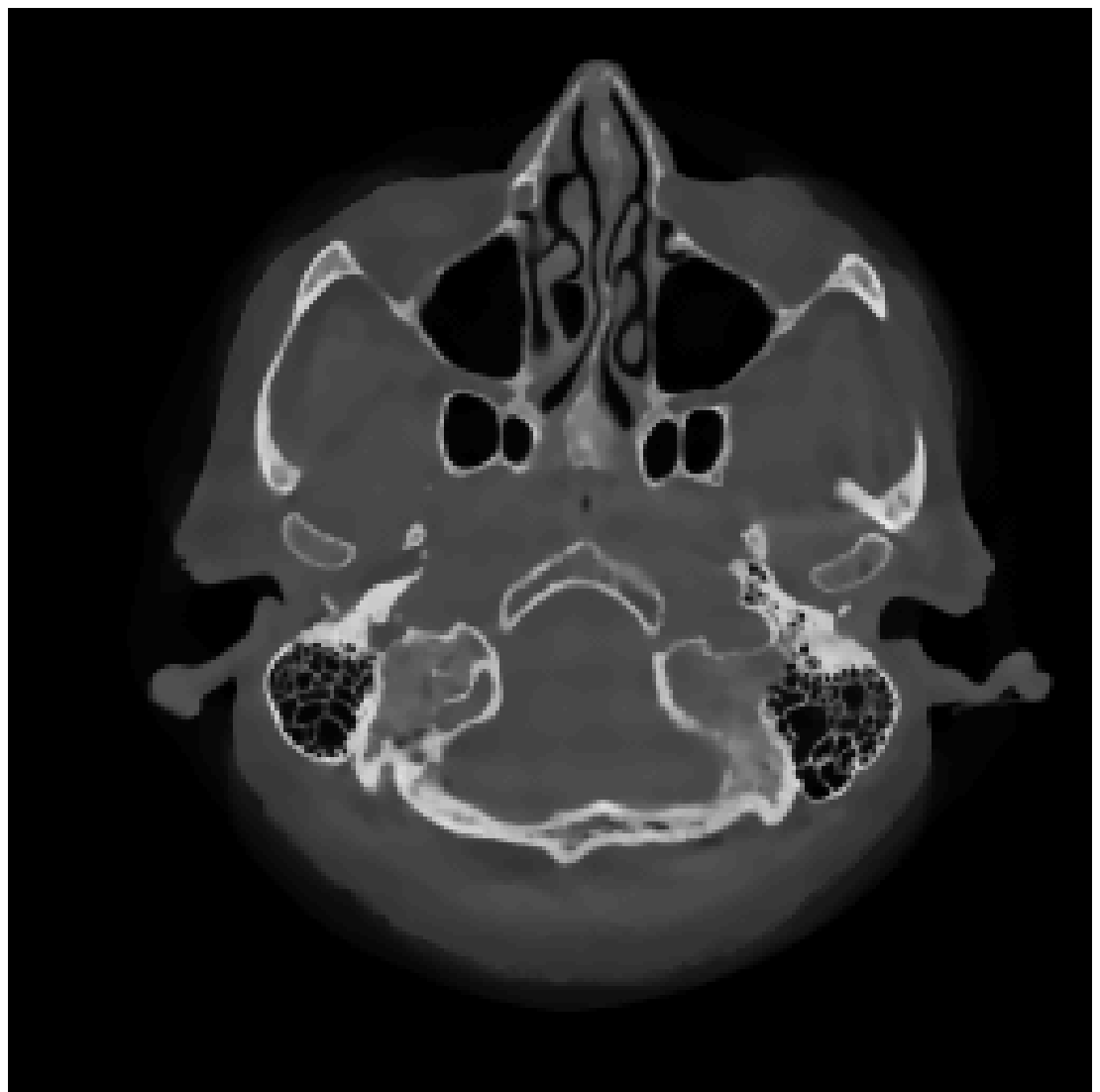}}
   \caption{Reconstruction results for the clinical data ($\kappa = 0.6p_{\max}$): (a) ground truth; (b) FBP-WCE; (c) SART-ISD; (d) M1bit-CSR-ISD.}\label{fig-real-result-15}
\end{figure}

\begin{figure}[htbp]
  \centering
  \subfigure[]{\label{fig-real-projection:a}
    \includegraphics[width=0.75\linewidth]{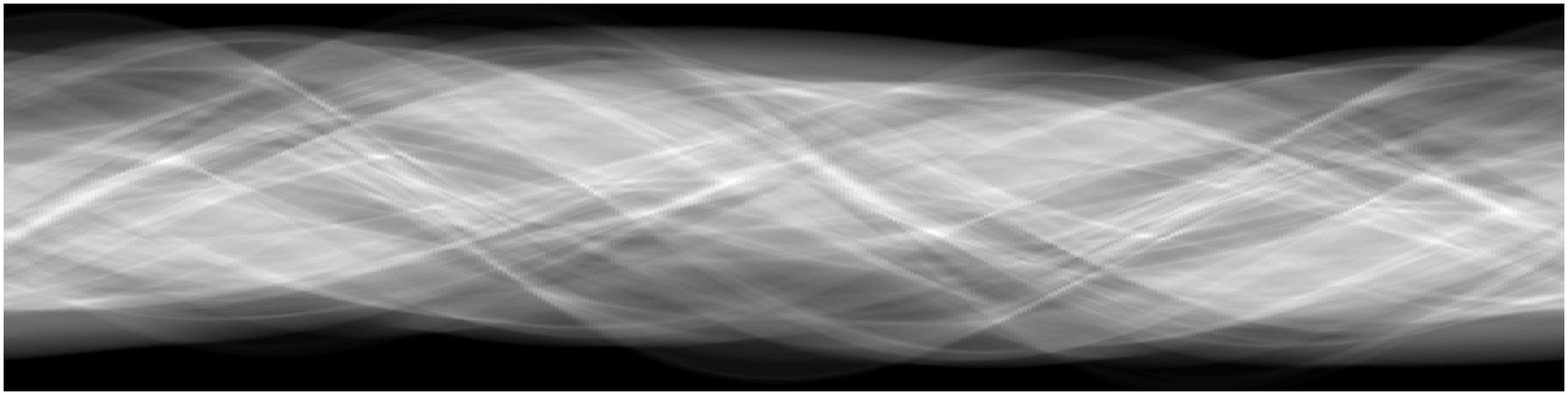}} \\
  \subfigure[]{\label{fig-real-projection:b}
    \includegraphics[width=0.75\linewidth]{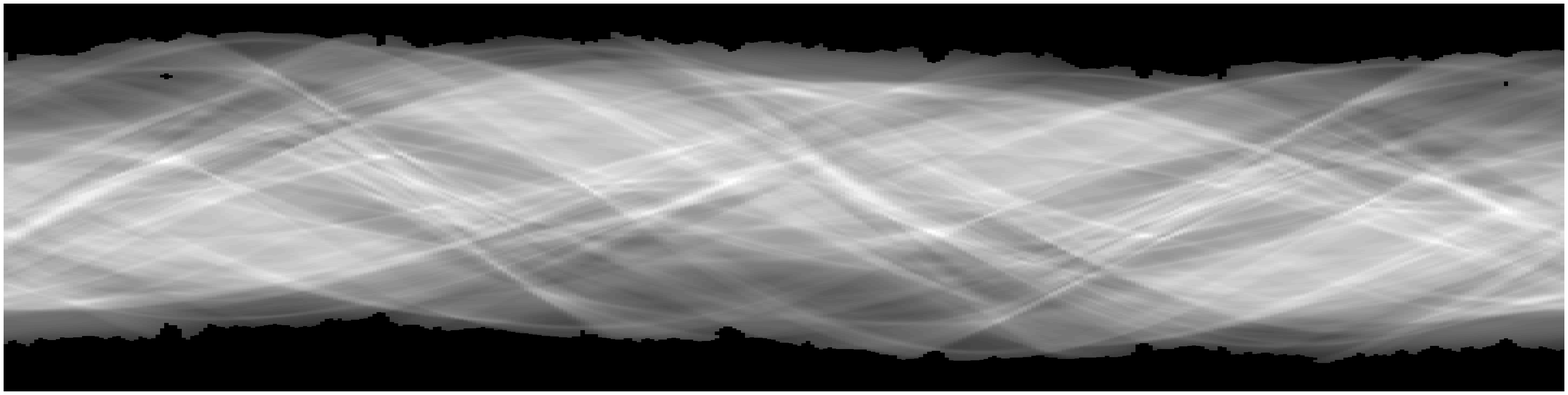}} \\
  \subfigure[]{\label{fig-real-projection:c}
    \includegraphics[width=0.75\linewidth]{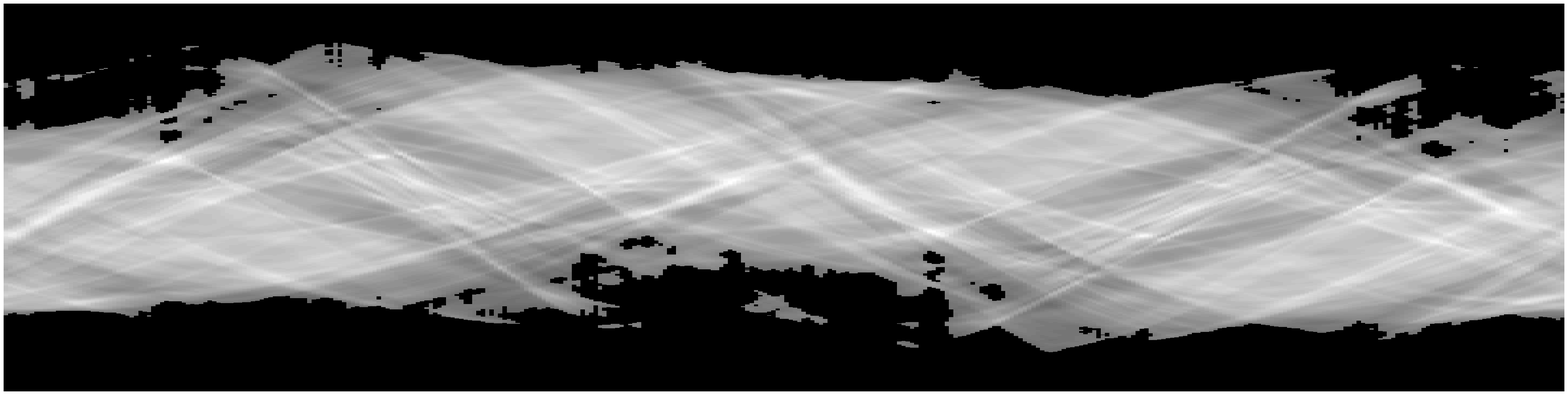}}
  \caption{Projections of Fig.\ref{fig-real-result-15:0}: (a) full projection data; (b) saturated projection data with $\kappa = 0.6p_{\max}$; (c) saturated projection data with $\kappa = 0.4p_{\max}$.}\label{fig-real-projection}
\end{figure}

For $\kappa = 0.6p_{\max}$, the reconstruction results of FBP-WCE and SART-ISD are given in Fig.\ref{fig-real-result-15:a} and \ref{fig-real-result-15:b}, respectively. As shown before, the traditional FBP method cannot handle the saturated data. With water cylinder extrapolation, the reconstruction quality has been improved but loss of clear patient boundaries still happens. The overall performance of SART-ISD is slightly better than FBP-WCE but capping artifact can be identified at the object border. Further improvement is obtained using the proposed M1Bit-CSR-ISD to acquire information from the saturated data. As shown in Fig.\ref{fig-real-result-15:c}, most of outer boundaries are nicely restored and streaking artifacts are effectively eliminated.


Next, we artificially add Gaussian noise on the projections and the  standard deviation of the noise $\sigma = 0.1$. The reconstructed images of FBP-WCE (with a smooth filter), SART-ISD, and M1bit-CSR-ISD are displayed in Fig.\ref{fig-real-result-noise}. Comparing the images in Fig.\ref{fig-real-result-15:c} and \ref{fig-real-result-noise:c}, we can find that the proposed method is not sensitive to additive noise, which is rooted in the loss functions of (\ref{M1bit-CS}) and (\ref{M1bit-CS-2}). If the projections contain outliers, then some robust losses can be applied.

\begin{figure}[htbp]
  \centering
  \subfigure[]{\label{fig-real-result-noise:0}
    \includegraphics[width=0.45\linewidth]{fig/real.eps}}\quad
  \subfigure[]{\label{fig-real-result-noise:a}
    \includegraphics[width=0.45\linewidth]{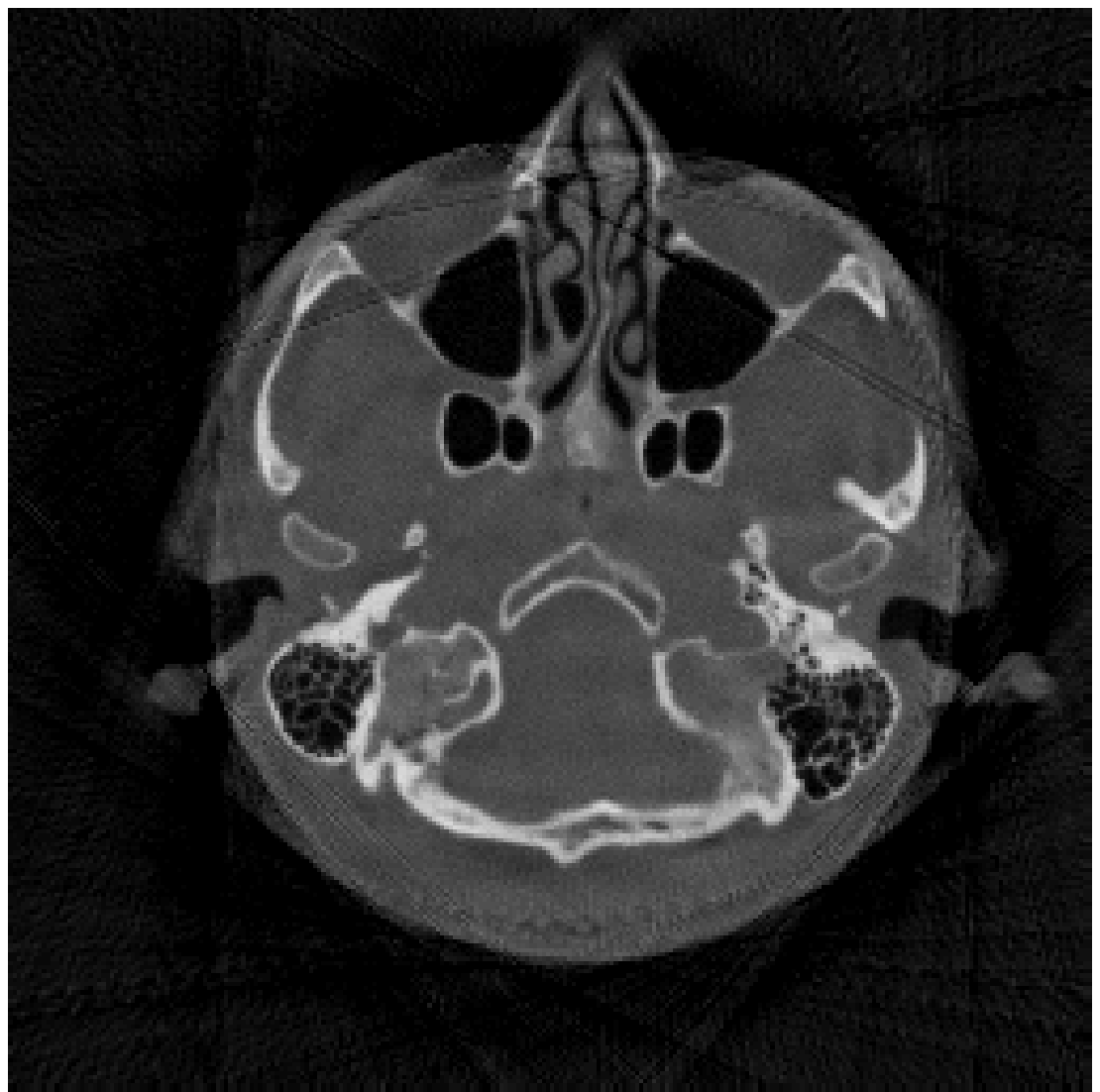}} \\
  \subfigure[]{\label{fig-real-result-noise:b}
    \includegraphics[width=0.45\linewidth]{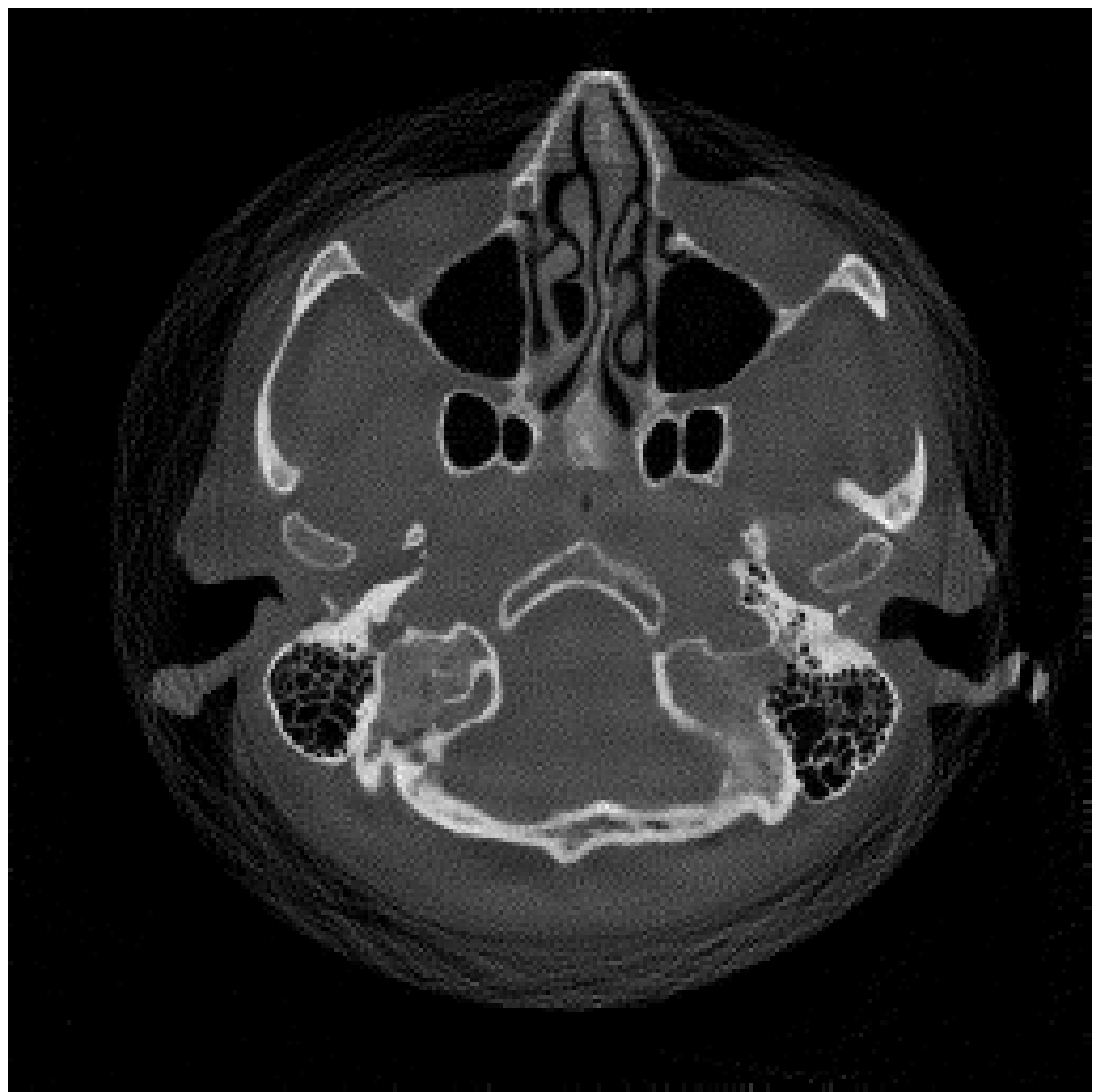}}\quad
  \subfigure[]{\label{fig-real-result-noise:c}
    \includegraphics[width=0.45\linewidth]{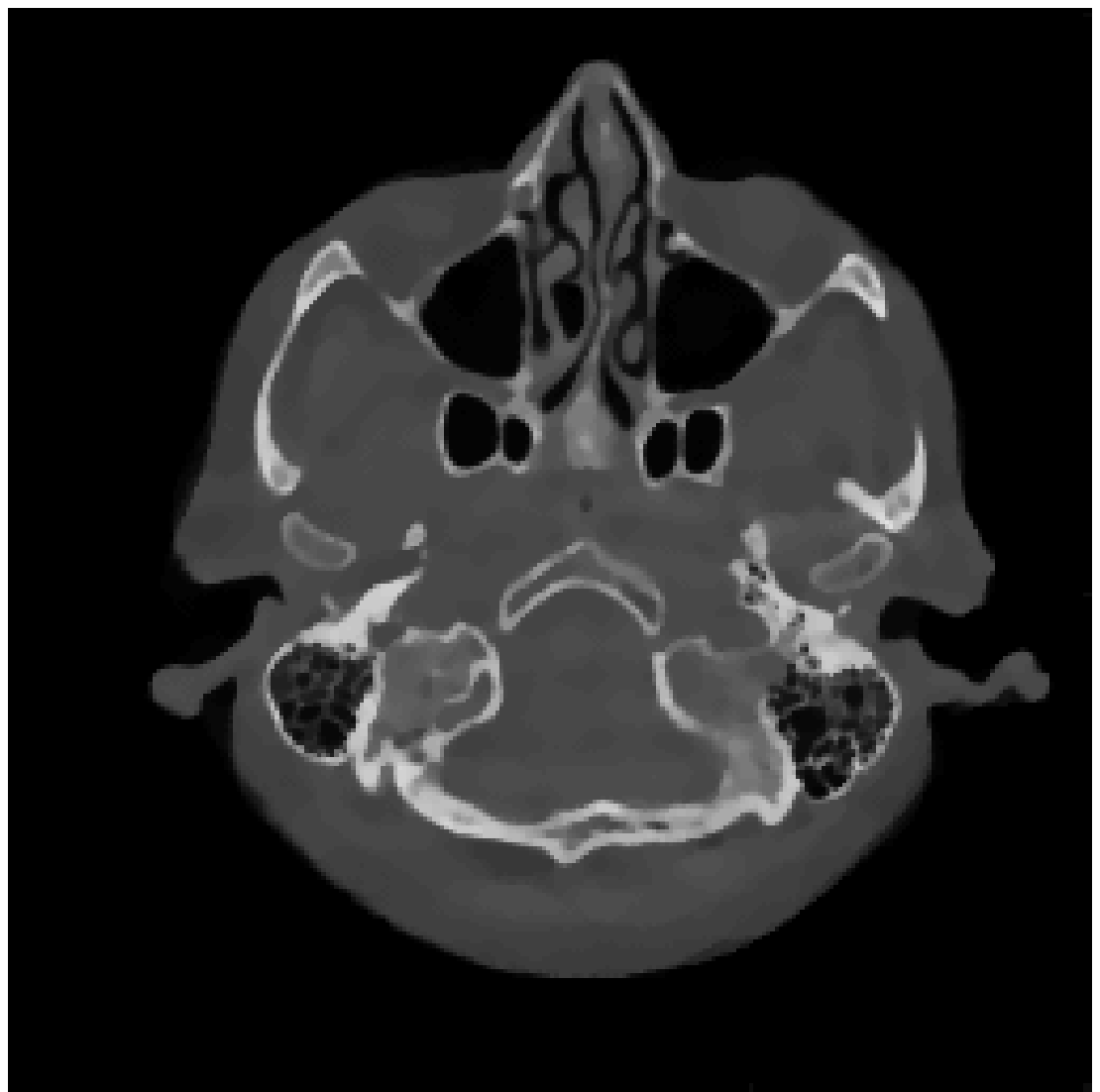}}
   \caption{Reconstruction results from the noise-corrupted and saturated projections ($\kappa = 0.6p_{\max}$ and $\sigma = 0.1$): (a) ground truth; (b) FBP-WCE; (c) SART-ISD; (d) M1bit-CSR-ISD.}\label{fig-real-result-noise}
\end{figure}

When the detectable range is smaller such that $\kappa = 0.4p_{\max}$, saturation becomes worse. Then the performance of FBP and SART, even with water cylinder extrapolation and the ISD scheme, dramatically drops. In practice, this heavy saturation rarely happens and is out of the scope of FBP-WCE. But even with this heavy saturation, the proposed M1bit-CSR-ISD can output a good result. The reconstructed images and enlargements of a region are illustrated in Fig.\ref{fig-real-result-10} and Fig.\ref{fig-real-result-10-part}, respectively. In this case, both FBP-WCE and SART-ISD fail to restore the clear outer boundaries of the patient, while M1bit-CSR-ISD is still able to achieve this in a more accurate manner. We further report the saturation detection result in Fig.\ref{fig-real-saturation_matrix-10}, from which one can observe that most of the saturations have been properly detected.

\begin{figure}[htbp]
  \centering
  \subfigure[]{\label{fig-real-result-10:0}
    \includegraphics[width=0.45\linewidth]{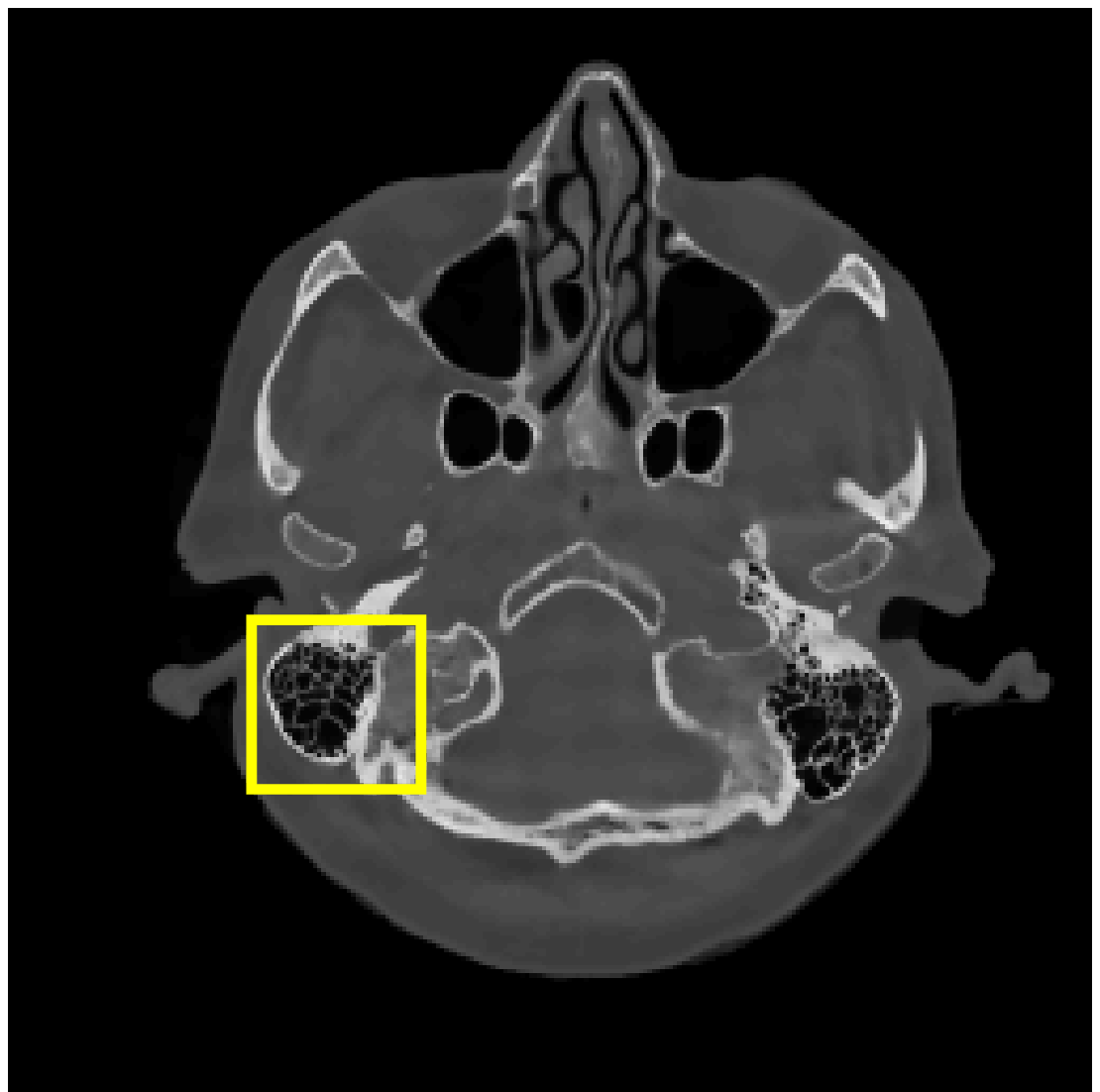}}\quad
  \subfigure[]{\label{fig-real-result-10:a}
    \includegraphics[width=0.45\linewidth]{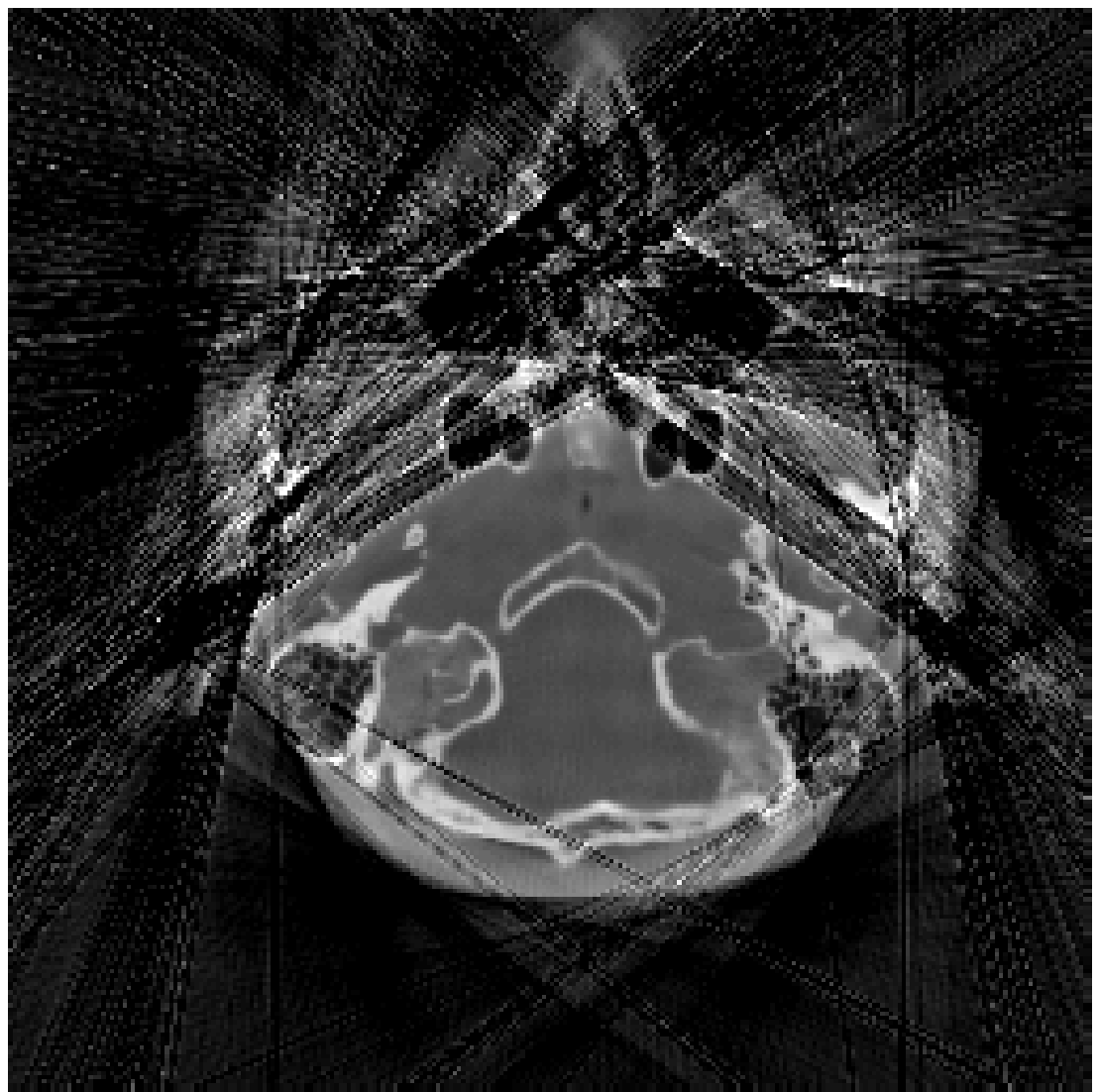}} \\
  \subfigure[]{\label{fig-real-result-10:b}
    \includegraphics[width=0.45\linewidth]{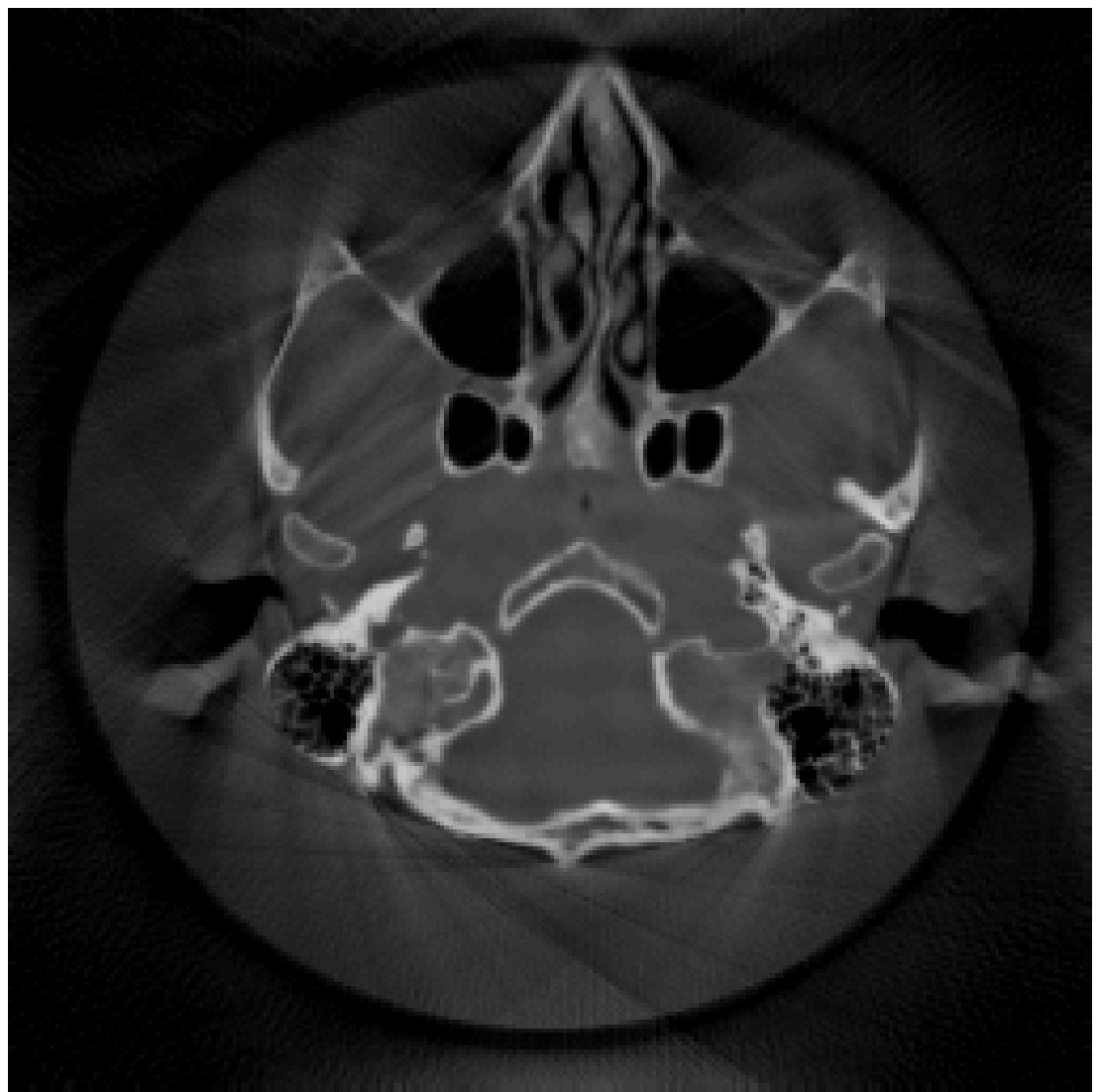}}\quad
  \subfigure[]{\label{fig-real-result-10:c}
    \includegraphics[width=0.45\linewidth]{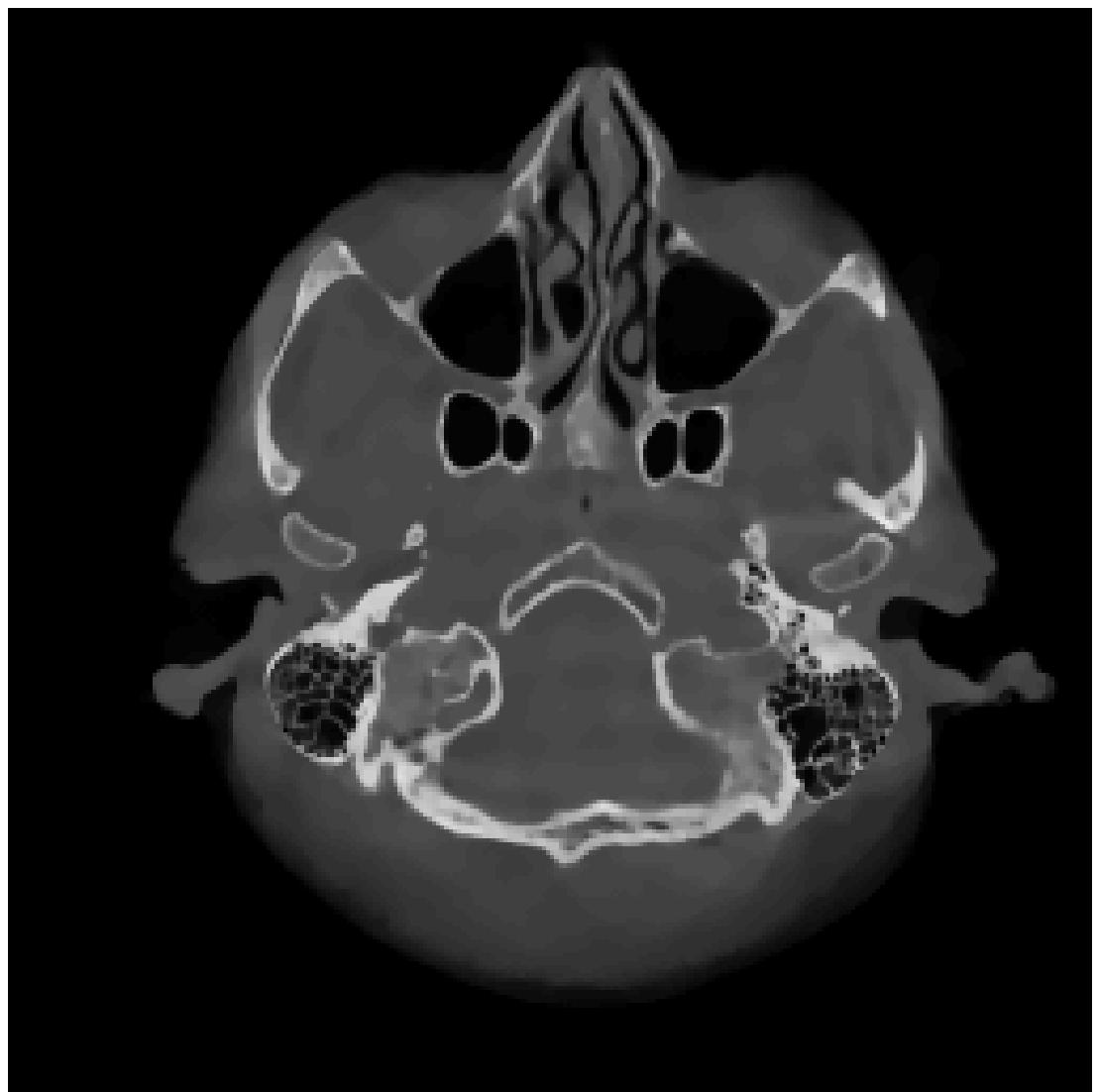}}
  \caption{Reconstruction results for the clinical data ($\kappa = 0.4p_{\max}$): (a) ground truth (yellow rectangle is enlarged in Fig.\ref{fig-real-result-10-part}); (b) FBP-WCE; (c) SART-ISD; (d) M1bit-CSR-ISD.}\label{fig-real-result-10}
\end{figure}

\begin{figure}[htbp]
  \centering
  \subfigure[]{\label{fig-real-result-10-part:0}
    \includegraphics[width=0.22\linewidth]{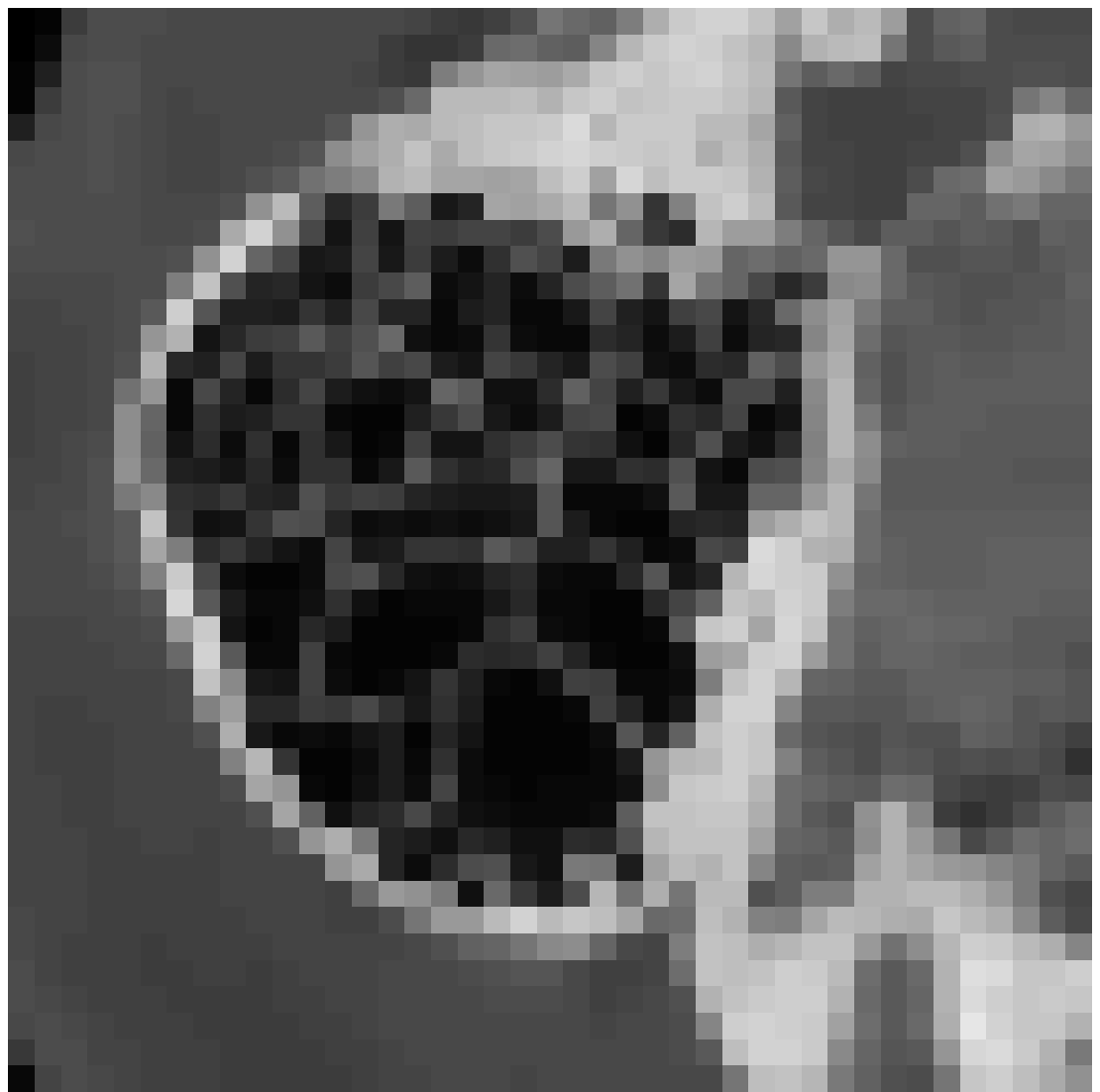}}
  \subfigure[]{\label{fig-real-result-10-part:a}
    \includegraphics[width=0.22\linewidth]{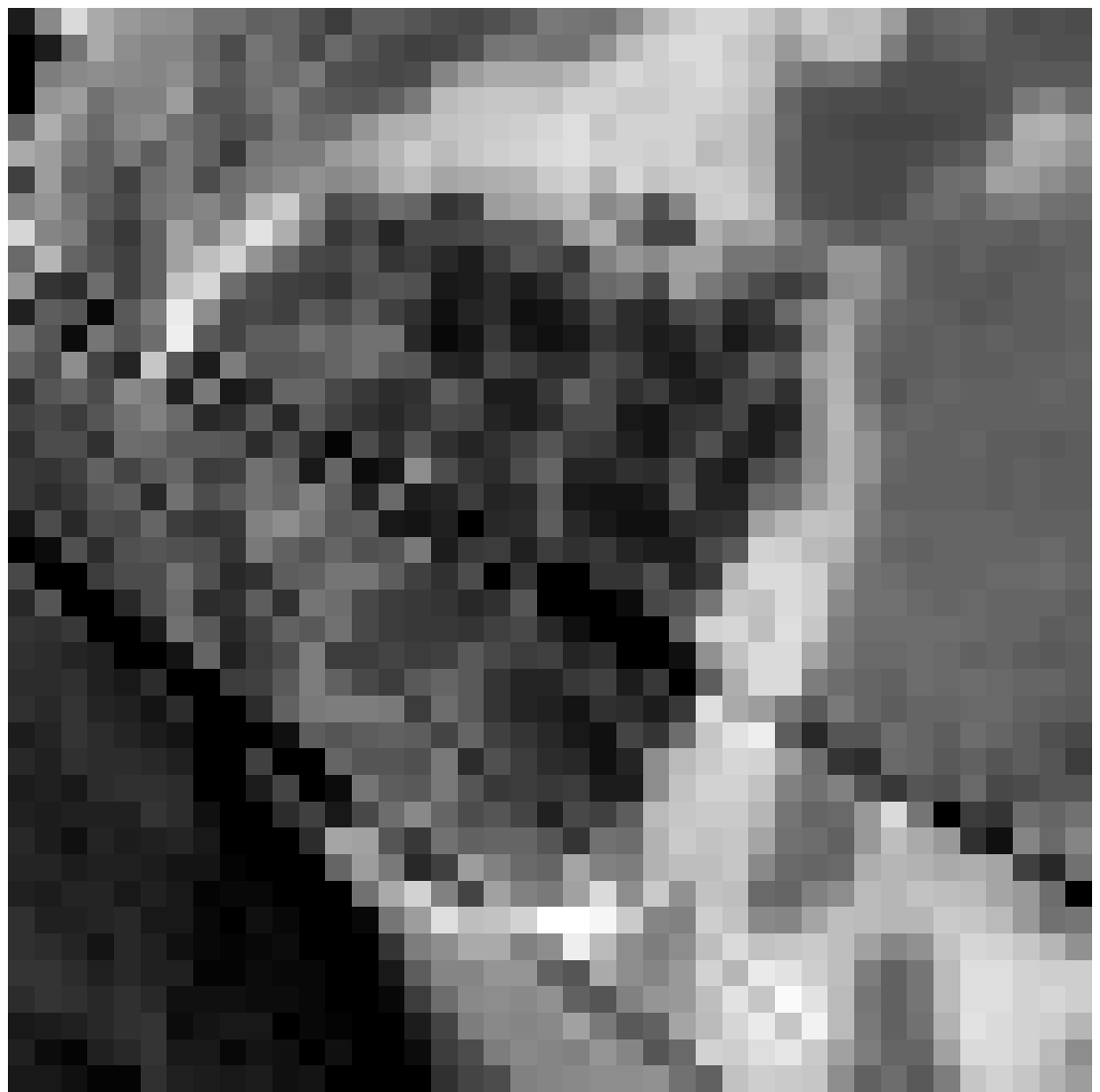}}
  \subfigure[]{\label{fig-real-result-10-part:b}
    \includegraphics[width=0.22\linewidth]{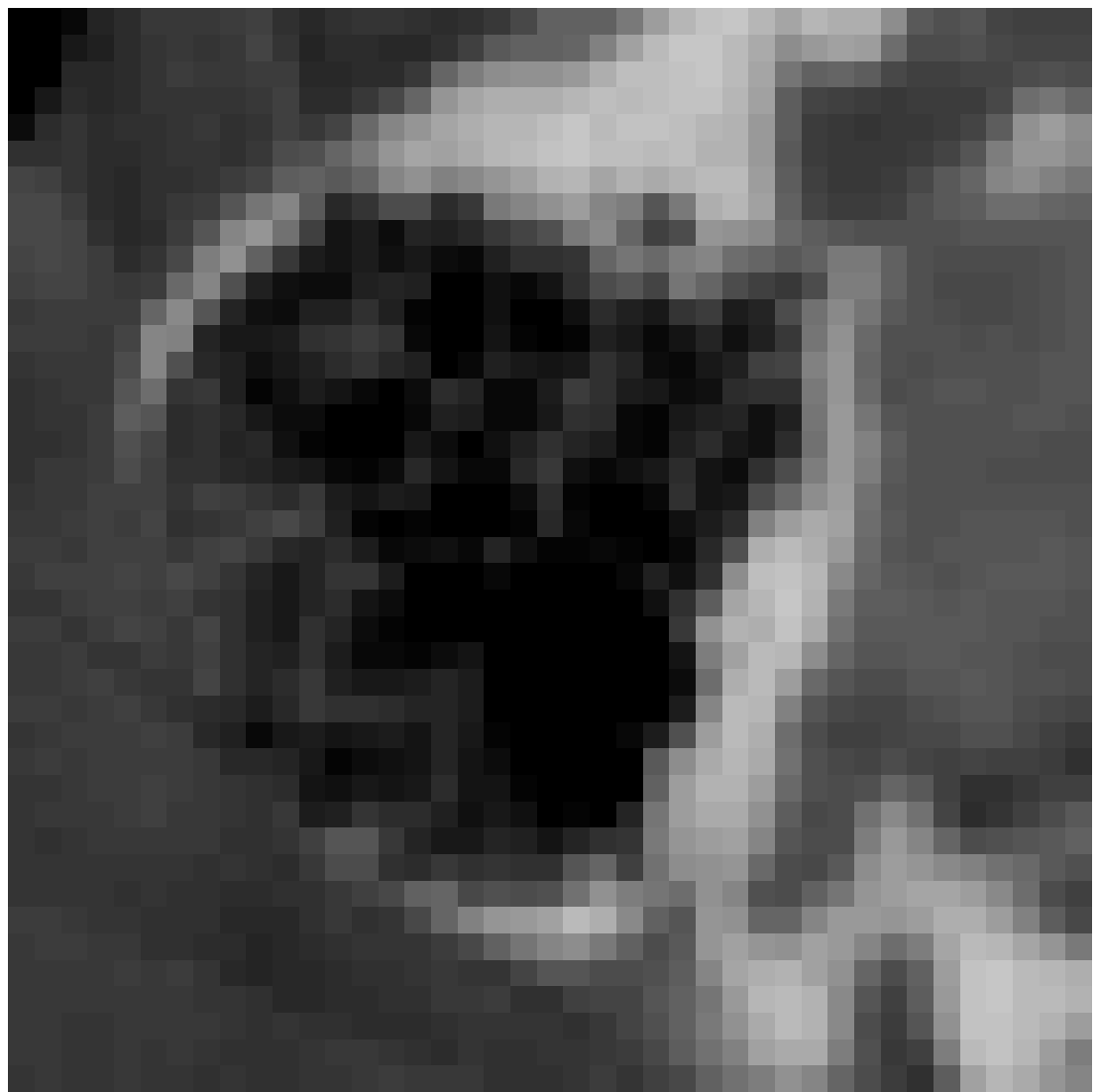}}
  \subfigure[]{\label{fig-real-result-10-part:c}
    \includegraphics[width=0.22\linewidth]{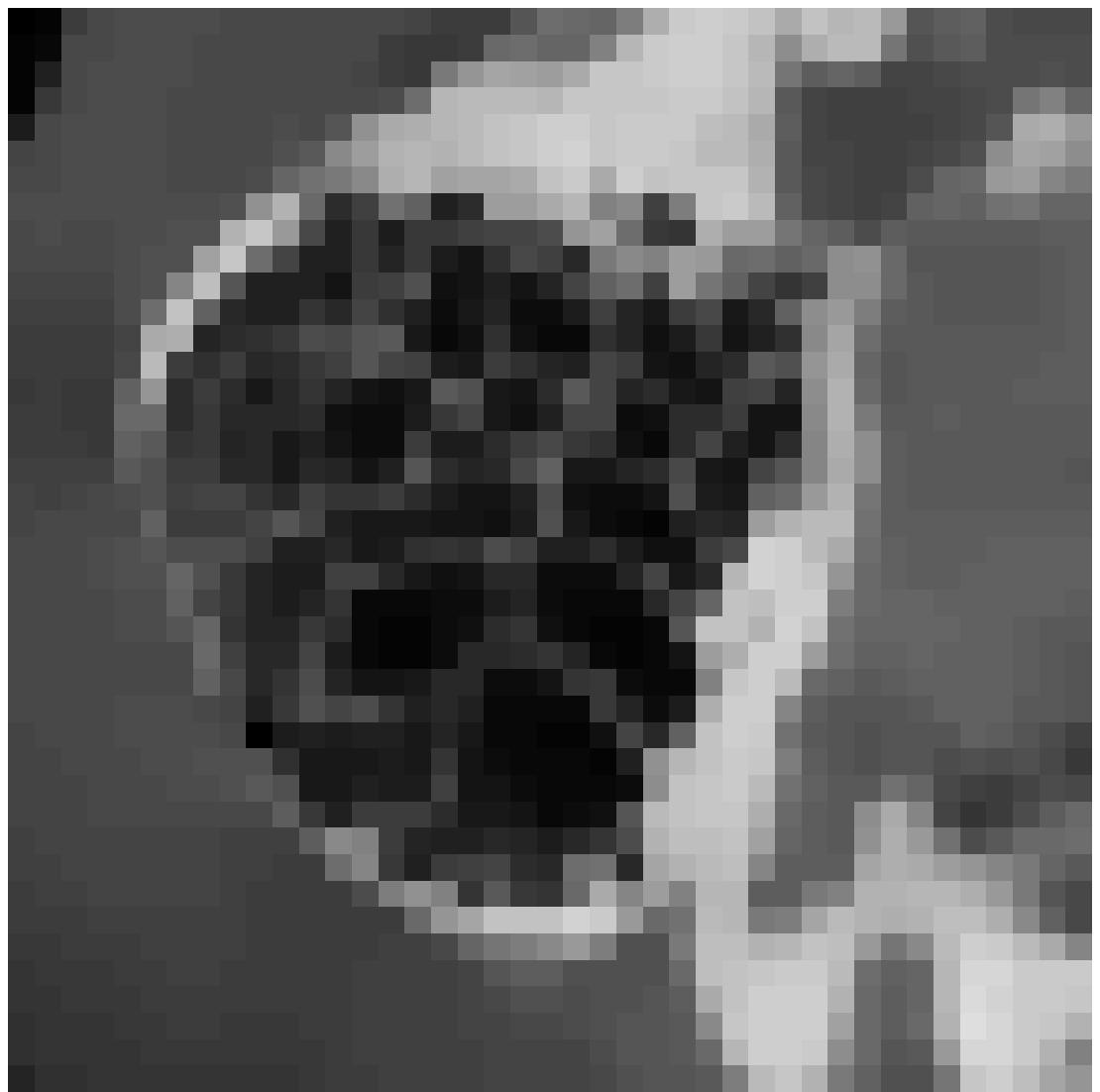}}
  \caption{An enlargement of a region of the clinical data, which is marked by yellow rectangle in Fig.\ref{fig-real-result-10:0}: (a) ground truth; (b) FBP-WCE; (c) SART-ISD; (d) M1bit-CSR-ISD.}\label{fig-real-result-10-part}
\end{figure}

\begin{figure}[htbp]
  \centering
  \subfigure[]{\label{fig-real-saturation_matrix-10:a}
    \includegraphics[width=0.75\linewidth]{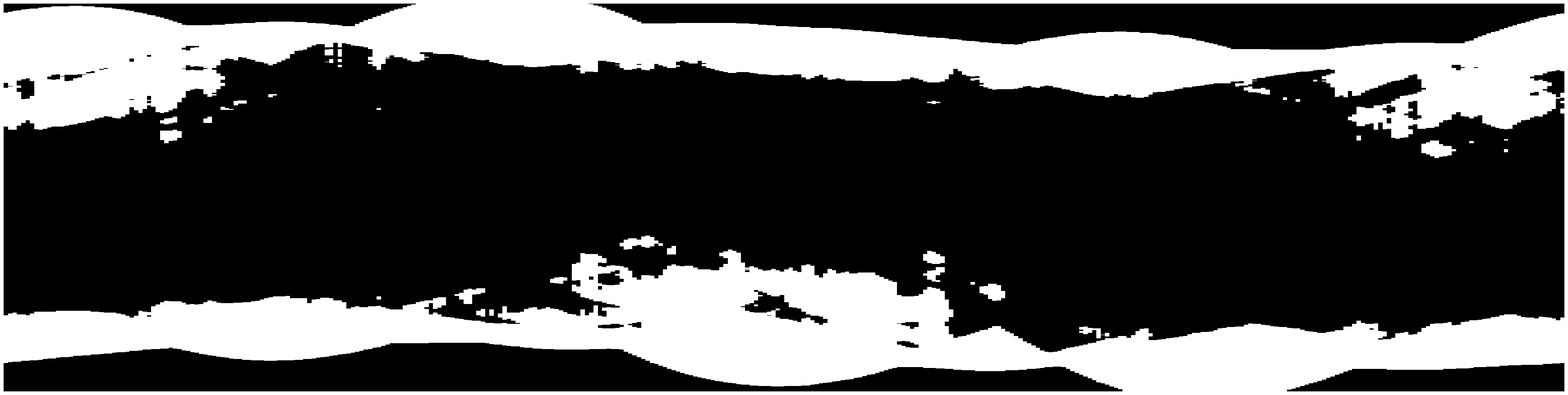}} \quad
  \subfigure[]{\label{fig-real-saturation_matrix-10:b}
    \includegraphics[width=0.75\linewidth]{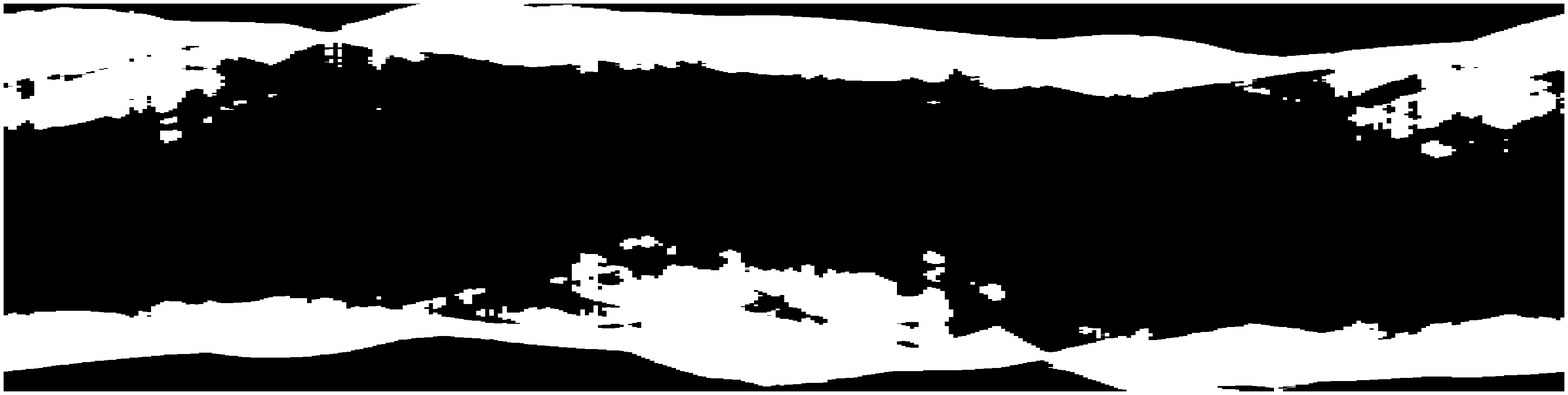}}
  \subfigure[]{\label{fig-real-saturation_matrix-10:b}
    \includegraphics[width=0.75\linewidth]{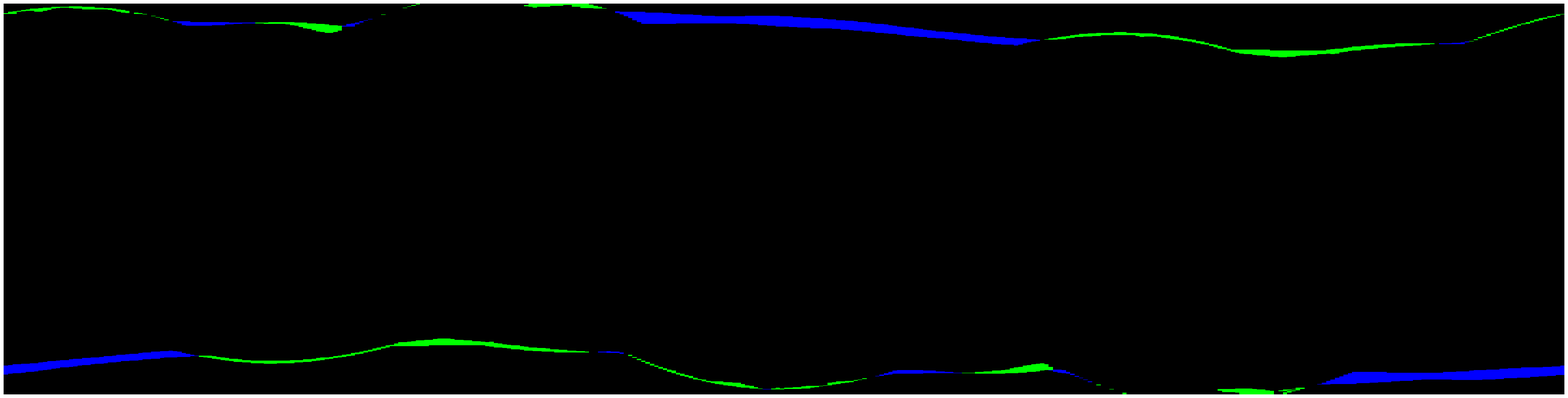}}
  \caption{Saturation indicator for the clinical phantom with $\kappa = 0.4p_{\max}$. (a) the true saturation indicator; (b) $\Psi$ detected by M1bit-CSR-ISD; (c) difference between the true and the detected saturation indicator (blue: false detection; green: missing detection).}\label{fig-real-saturation_matrix-10}
\end{figure}


In Table \ref{table-err}, we list the RMSE (in HU) for FBP, FBP-WCE, SART-ISD, M1bit-CSR with ISD and with an ideal saturation matrix. Compared with classical FBP, FBP-WCE and SART-ISD have significant improvements. But the schemes are basically different: FBP-WCE applies water cylinder extrapolation as the complements of the missing projections, and SART-ISD is to iteratively detect and drop incorrect projections. The proposed M1bit-CSR can adequately use the information carried by overexposed measurements and then outputs good reconstruction results. The gap between M1bit-CSR-ISD and M1bit-CSR with the ideal $\Psi$ is not very big, which on the one hand shows the accuracy of the proposed ISD method, and on the other hand confirms the robustness of M1bit-CSR against a part of incorrect detection.

\begin{table}[htbp]
\centering
\caption{\label{table-err} Root of Mean Square Error (HU) of Reconstruction Results}
\scriptsize
\begin{tabular}{@{}l|cccccc}
\toprule
   & FBP & FBP & SART & M1bit-CSR & M1bit-CSR \\
 data & & WCE & ISD & ISD & (ideal $\Psi$)\\
\midrule
knee phantom  \\
($\kappa = 0.5p_{\max}$) & 182.4 & 97.93 & 54.91 & 8.547 & 5.461 \\
head     \\
($\kappa = 0.6p_{\max}$) & 43.22 & 35.34 & 31.97 & 11.62 & 6.930 \\
head with noise    \\
($\kappa = 0.6p_{\max}$) & 43.84 & 36.71 & 34.22 & 11.72 & 7.001 \\
head     \\
($\kappa = 0.4p_{\max}$) & 132.2 & 87.93 & 70.63 & 14.81 & 12.39\\
\bottomrule
\end{tabular}
\end{table}


In M1bit-CS models, the saturation indicator $\Psi$ is a key input. In this paper, we designed a simple but efficient iterative method to detect $\Psi$. Further improvement may come from the use of prior-knowledge and reasonable assumptions on tissue structures. At least, those prior-knowledge provides a good initial guess. An accurate initial guess can reduce both the reconstruction error and the detection time.

\section{Conclusion}
\label{sec:conclusion}

Aiming at signal reconstruction when saturation happens, we established mixed one-bit compressive sensing that could be regarded as a bridge between the regular and one-bit CS. We developed the corresponding ADMM and an iterative saturation detection method. They are then successfully applied to overexposure correction for C-arm CT reconstruction. The improvement from the existing methods is quite significant. Generally, the reconstruction performance of the proposed method is satisfactory even in the presence of severe detector saturation, showing the benefit of our method on considering saturated projections.

CT scanning that has overexposure is a typical example of sensing systems with saturation. The promising performance of M1bit-CS implies its potential use on other tasks with measurements beyond the reliable sensing region. We can even take advantage of saturation phenomena for specific purposes. For example, one can sample signs for a part of the measurements to reduce the expense of the analog-to-digital conversion without significantly sacrificing the recovery accuracy. In low-dose computed tomography, when we largely reduce the radiation dose to the level such that some projections are below the threshold, then those projection values become unreliable but they do provide one-bit information. Then M1bit-CS can also be applied to improve the performance.

\section*{Acknowledgment}

The authors would like to thank Prof. Ji Liu from the University of Rochester for sharing the code of RDCS.


%





\bibliographystyle{IEEEtran}        
\bibliography{refs}


\end{document}